\documentclass[10pt,twocolumn,letterpaper]{article}
\usepackage{xcolor,pifont}
\usepackage{comment}
\usepackage{stfloats}

 \usepackage{cvpr}              % To produce the CAMERA-READY version
\usepackage{float}
\definecolor{cvprblue}{rgb}{0.21,0.49,0.74}
\usepackage[pagebackref,breaklinks,colorlinks,allcolors=cvprblue]{hyperref}

\def\paperID{10} % *** Enter the Paper ID here
\def\confName{CVPR}
\def\confYear{2026}

\title{EgoCampus: Egocentric Pedestrian Eye Gaze Model and Dataset}

\author{Ronan John$^\dag$ \\
\and
Aditya Kesari$^\dag$ \\
\and
Vincenzo DiMatteo \\
\and
Kristin Dana \\
\and
 Rutgers University, New Brunswick
\and
\\
{\tt\small ronan.john, aditya.kesari, vincenzo.dimatteo, kristin.dana@rutgers.edu}
}

\begin{document}
\maketitle
\begin{abstract}
We address the challenge of predicting human visual attention during real-world navigation by measuring and modeling egocentric pedestrian eye gaze in an outdoor campus setting. 
We introduce the \textbf{EgoCampus} dataset, which spans 25 unique outdoor paths over 6 km across a university campus with recordings from more than 80 distinct human pedestrians, resulting in a diverse set of gaze-annotated videos. The system used for collection, Meta's Project Aria glasses, integrates eye tracking, front-facing RGB cameras, inertial sensors, and GPS to provide rich data from the human perspective. Unlike many prior egocentric datasets that focus on indoor tasks or exclude eye gaze information, our work emphasizes visual attention while subjects walk in outdoor campus paths. Using this data, we develop \textbf{EgoCampusNet}, a novel method to predict eye gaze of navigating pedestrians as they move through outdoor environments. Our contributions provide both a new resource for studying real-world attention and a resource for future work in gaze prediction models for navigation. Dataset and code will be made publicly available at a later date at \url{https://github.com/ComputerVisionRutgers/EgoCampus}.
\end{abstract}

\def\thefootnote{\dag}\footnotetext{Denotes equal contribution}\def\thefootnote{\arabic{footnote}}
\section{Introduction}
\label{sec:intro}

\begin{figure}[t!]
\centering
\includegraphics[width=0.95\linewidth]{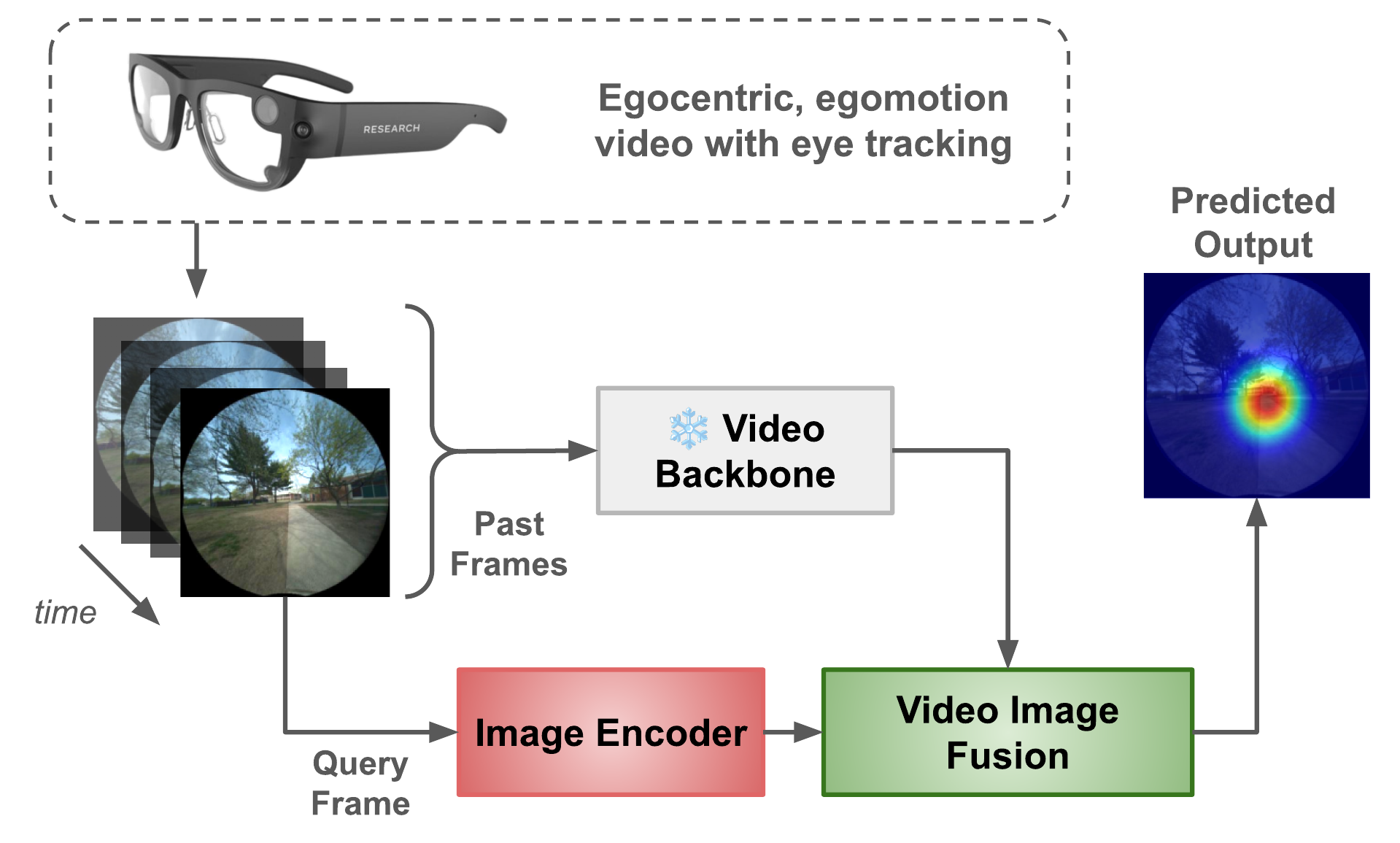}
\caption{\textbf{Overview of our proposed gaze prediction model, ECN}. The Project Aria glasses capture a stream of egocentric video. From this video, we extract features with a pretrained backbone. From the query image (usually the last frame), we extract features with a trained image encoder. Lastly, the video and image features are fused and decoded in order to predict the final output, representing where people are most likely to look during egocentric motion.}
\label{fig:teaser}
\end{figure}

Understanding where humans look as they navigate the world is fundamental for interpreting behavior and intent. Gaze patterns reveal common fixations, saccades, and attention shifts, offering insight into how people prioritize visual information in complex environments. Understanding human gaze and visual attention is particularly relevant for training embodied agents operating in shared human spaces,  where  prediction models may support navigation, cooperation, or anticipatory behavior. 
Much of the prior research in eye gaze has focused on constrained tasks or environments, such as gaze on static images, screen-based videos, structured indoor activities like cooking, cleaning, and carpentry. 
The recent large-scale egocentric dataset Ego4D \cite{grauman2024ego4d} is a pioneering dataset with large  variation of subjects, tasks and context. While Ego4D primarily consists of egocentric videos of tasks,  
eye gaze was captured on a small subset and the focus is  not outdoor pedestrian locomotion.
 
Research on eye gaze in outdoor environments during navigation is still in its early stages.

\begin{figure*}[t!]
\centering
\includegraphics[width=1.0\linewidth]{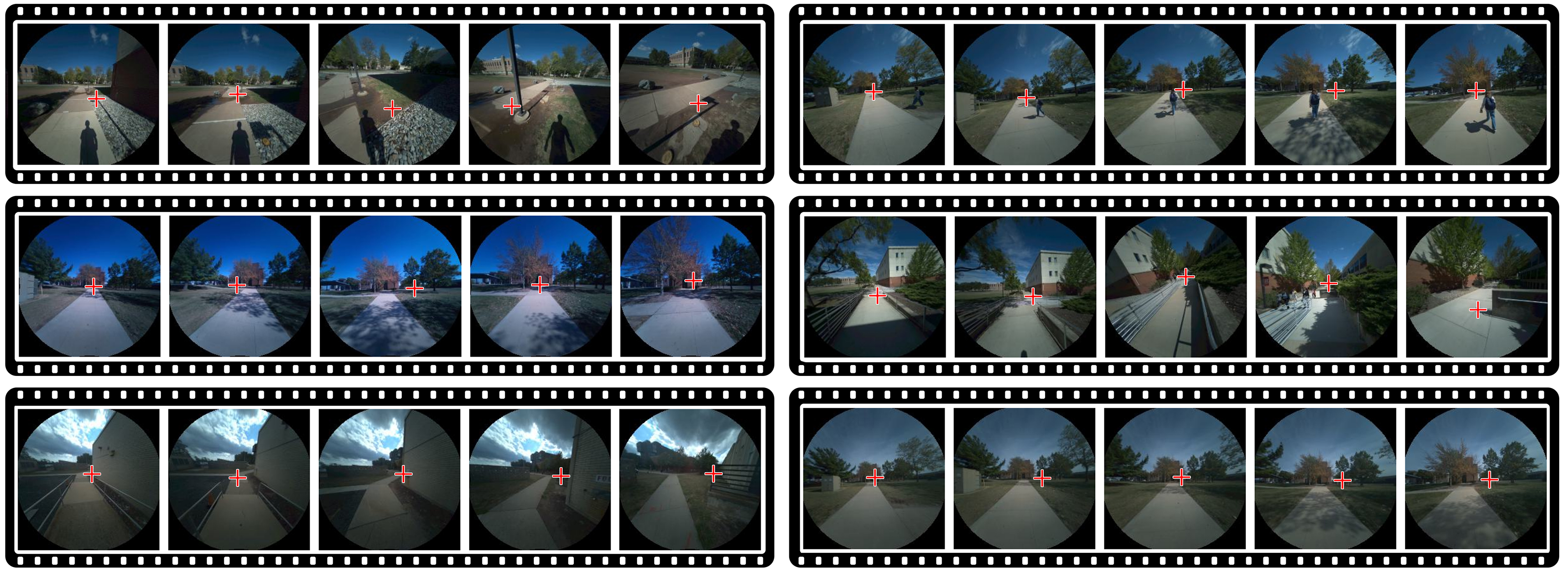}
\caption{\textbf{Samples of video sequences from the EgoCampus dataset.} EgoCampus contains 32 hours of video from the Project Aria glasses, capturing egocentric video from 82 pedestrians traversing 25 distinct paths on a university campus. Each frame has an associated eye gaze coordinate (shown above as the red ``+'') and auxiliary sensor readings (IMU, GPS,  Wi-Fi). }
\label{fig:dataset_sample}
\end{figure*}

% what are we doing to address this gap?
In this work, we introduce: (1)  EgoCampus, a pedestrian egocentric dataset capturing synchronized eye gaze, video, and sensor data as participants navigate real-world outdoor routes on a university campus; and (2) EgoCampusNet \textbf{(ECN)} (\cref{fig:teaser}), a prediction model leveraging a pre-trained video backbone and an eye gaze predictor head. 
%Our dataset provides structured navigation paths with multiple participants traversing the same routes. 
EgoCampus is tightly focused on  eye gaze of pedestrians walking outdoors in a campus setting, and it comprises multiple subjects traveling similar routes. This framework enables both individual analysis and cross-subject comparisons of gaze behavior across shared spatio-temporal contexts. All data is collected using Meta's Project Aria glasses~\cite{engel2023projectarianewtool}, providing high-resolution RGB video, gaze coordinates, inertial measurements, and GPS. This combination allows us to model and analyze pedestrian eye gaze during locomotion over a range of subjects. Our EgoCampus dataset spans 25 unique outdoor paths over 6 km across a university campus with recordings from more than 80 distinct human pedestrians (see ~\cref{fig:dataset_sample}). The entire dataset will be made publicly available and provides a valuable resource for understanding pedestrian spatial attention, with strong potential for navigation applications such as developing spatio-temporal environment-sampling methods that mimic human attention.

% how do we evaluate and validate our results? note: cant be finalized till model is finalized
% From Kristin: I took this paragraph out, it wasn't updated in a long tie and no longer fits. It need stay removed or be rewritten. Perhaps some can go in related work
\begin{comment}
We explore the gaze prediction task to model and analyze this behavior. Gaze maps provide a compact visual representation of where conspicuity is to occur, highlighting prominent regions of fixation \cite{yamada_2010_SaliencyPredictGaze}. These maps have long been used in computer vision to represent gaze density in images or video, whether derived from empirical human data or predicted by models such as TempSAL \cite{aydemir2023tempsal}, SUM \cite{Hosseini_2025_WACV}, EML-NET \cite{JIA20EML}, and CV\_MM \cite{moskalenko2024aim} to name a few. Recent advances in deep learning, particularly convolutional neural networks (CNNs) \cite{kummerer2016deepgaze} and transformer-based architectures \cite{vaswani2023attentionneed}, have significantly improved the ability to predict saliency. While some models operate on static images, others integrate temporal information from videos, making them suitable for egocentric video saliency tasks.
\end{comment}

% what is the novelty in our contributions? + summary of intro
To summarize, our contributions are as follows:
\begin{itemize}
    \item An egocentric video dataset (with eye gaze) collected during pedestrian locomotion. (EgoCampus)
    \item A novel model for efficiently learning to predict gaze from egocentric video. (EgoCampusNet)
    \item An evaluation of model performance using the EgoCampus dataset including a novel strategy for weighting metrics relative to data prior
\end{itemize}

\section{Related Work}
%
%
% if we need more space, replace \subection with \bold
\paragraph{\bf Gaze Prediction.}
As a person moves through their environment, their attention is often split between objects that are visually salient and objects that provide task-relevant information \cite{corbetta2002control,wolfe2017five}. Gaze prediction requires task-specific dependencies to be captured in order to accurately predict fixations. Li et al. \cite{Li_2013_ICCV} sought to predict eye gaze in egocentric video, modeling the temporal dynamics of gaze for object segmentation and action recognition. They advanced this research by combining head motion in gaze prediction in egocentric settings.
Seminal work \cite{fathi2012learning} established that gaze and actions are mutually informative in egocentric video in the context of recognition. 
The scope was expanded beyond ``where people look'' to ``when they stop looking'', modeling human attention dynamics even when visual goals are not present \cite{yang2022target}. Park et al. \cite{Park_2015_CVPR} modeled saliency in social settings, encoding the relative positions of people in a scene to capture task-specific interaction. Work from Thakur et al. \cite{thakur_social_imu_2021} also predicted gaze in a social setting, but incorporated IMU readings from the capture device. More modern work \cite{mondal2023gazeformer, cartella2025modeling, Lai2024}  integrates transformers and diffusion models for efficient goal-directed gaze prediction.

\paragraph{\bf Saliency Prediction.}
Saliency maps offer a compact, interpretable representation of where attention concentrates, making them well suited for modeling egocentric video data. Early computational models, such as the Itti–Koch multi‑scale feature integration approach \cite{730558}, established the principle that visual contrasts drive attention. Graph‑Based Visual Saliency (GBVS) \cite{harel2006graph} further demonstrated how global graph normalization yields robust saliency. With deep learning, Convolutional Neural Networks (CNNs) dramatically improved prediction quality: Huang et al.\ \cite{7410395} introduced a dual‑stream CNN capturing coarse and fine features, and Jia et al.\ \cite{JIA20EML} proposed EML‑Net, an ensemble of pretrained backbones fused through multi‑layer decoding. Transformer architectures now dominate the field by modeling long‑range spatial and temporal dependencies. SUM \cite{Hosseini_2025_WACV} unifies image and video saliency with Mamba blocks, while recent works \cite{Lai2024,kumar2023eye} leverage token‑level attention across frames. Other recent advances include models that use pretrained backbones to assist in predicting saliency \cite{Linardos_2021_ICCV, kummerer2016deepgaze}, models that leverage saliency over time \cite{aydemir2023tempsal, Fosco_2020_CVPR}, and new saliency challenges \cite{moskalenko2024aim}.
%TempSal and Multi-Duration Saliency Excited Model which leverages saliency over the temporal space \cite{aydemir2023tempsal, Fosco_2020_CVPR} and benchmarks from the AIM Video Saliency Challenge \cite{moskalenko2024aim}.
A key line of research directly compares predicted saliency with recorded gaze, showing saliency maps correlate with human fixations in egocentric video \cite{yamada_2010_SaliencyPredictGaze}. 
However, most evaluations remain confined to static images or short clips.
We compare saliency and measured eye gaze by aggregating gaze across continuous outdoor traversals (after aligning frames). In this manner, we introduce a new setting for assessing how modern saliency models align with true spatial attention during navigation, advancing beyond isolated frame‑level benchmarks to cumulative, environment‑aware attention modeling in the wild.
\paragraph{\bf Egocentric and Gaze Prediction Datasets.}
Egocentric datasets study how individuals interact with the world by recording first-person viewpoint video from an egocentric camera. 
Studying eye movements to record gaze was a complex task in the past, with in-lab apparatus required \cite{kowler2011eye}.
Traditionally, eye gaze datasets were obtained with human subjects in a lab looking at a screen observing either static images or video. Examples of static-image studies include MIT1003 \cite{Judd_2009}, that uses eye-tracking to measure free-viewing saliency on desktop displays; SALICON \cite{Jiang_2015_CVPR}, which 
uses mouse movement as a proxy for eye-tracking to record saliency and scales via crowdsourced annotations; and Coco-search18 \cite{chen2021coco}, which has eye gaze from 10 people searching for each of 18 target-object categories in 6000+ natural-scene images.
% Another category for datasets is eye gaze while watching videos 
%often target specific problems: action recognition, and large scale image saliency. 
In video-based studies, the subject is fixed and watching pre-recorded actions \cite{wang2019neuro,wang2018revisiting,park2020towards}. In another paradigm called third-person gaze, the subjects are people depicted in static pre-captured images, and their static eye gaze direction in screen coordinates is estimated \cite{Kellnhofer_2019_ICCV,recasens2015they}. 
Prior paradigms with fixed scenes and static subjects are quite different from our EgoCampus scenario, where pedestrians walk and navigate in the real world. 
%EgoCampus measures eye gaze in-the-wild on long, repeatable outdoor navigation routes with synchronized multimodal sensing.

Advances in eye-tracking algorithms \cite{Krafka_2016_CVPR} paved the way for wearable headsets and glasses for mobile, in-the-wild eye gaze datasets. Similarly, the ubiquity of head-mounted cameras (e.g.\ GoPro) support egocentric datasets without eye gaze. 
Many of the recent datasets emphasize task completion in structured indoor scenes, such as EPIC-Kitchens \cite{Damen2018EPICKITCHENS}, HD-EPIC \cite{perrett2025hdepic}, HoloAssist \cite{wang2023holoassistegocentrichumaninteraction}, EgoExo4D \cite{grauman2024ego}, EgoExoLearn \cite{huang2024egoexolearn} and EGTEA Gaze+ \cite{li2021eye}, which provide gaze-annotated video for action recognition, primarily in kitchen environments. These datasets are not directly relevant to outdoor navigation.  Recent work on large-scale egocentric videos like Ego4D \cite{grauman2024ego4d} have limited eye-gaze measurements, especially for outdoor navigation. The GEETUP \cite{10.1145/3379156.3391378} dataset is more closely related to our work. This dataset depicts subjects navigating two paths utilizing the Tobii PRO Glasses 2 wearable eye trackers. The dataset provides RGB images, segmentation and depth maps with 43 subjects on two non-overlapping paths. 
Our dataset offers several advantages over GEETUP. GEETUP’s clips are segmented due to instances such as subjects pausing to talk or check phones, resulting in a median clip length of 14.7 seconds. In contrast, EgoCampus contains continuous path trajectories averaging 108.2 seconds per clip. EgoCampus also includes over twice as many subjects (82 pedestrians) and paths are traversed in both directions (forward and reverse). Additionally, the use of Project Aria glasses \cite{engel2023projectarianewtool} enables the capture of rich auxiliary data, in addition to egocentric eye gaze and video, including IMUs, barometer, magnetometer, GPS, and Wi-Fi/Bluetooth signals. 
After release of eye tracking glasses from Project Aria, 
 there have been recent expansions in egocentric datasets in varied environments \cite{Pan_2023_ICCV, kong2025ariagen2pilot, wang2025seeingdarkbenchmarkingegocentric, pan2025lookoutrealworldhumanoidegocentric}. Aria Navigation Dataset (AND) \cite{pan2025lookoutrealworldhumanoidegocentric} explores outdoor egomotion and develops a model for head pose, but the dataset seems to have few subjects (an exact count is not provided in the paper). 
As shown in ~\cref{tab:egodatasets} our EgoCampus dataset fills a gap in the current work, providing consistent egocentric RGB video and gaze recordings (augmented with IMU, GPS, Wi-fi) across a large number of  participants traversing identical outdoor campus routes, enabling cross-subject comparison in real-world navigation contexts.
\section{Dataset}\label{sec:dataset}

\begin{figure*}[t!]
\centering
\includegraphics[width=0.9\linewidth]{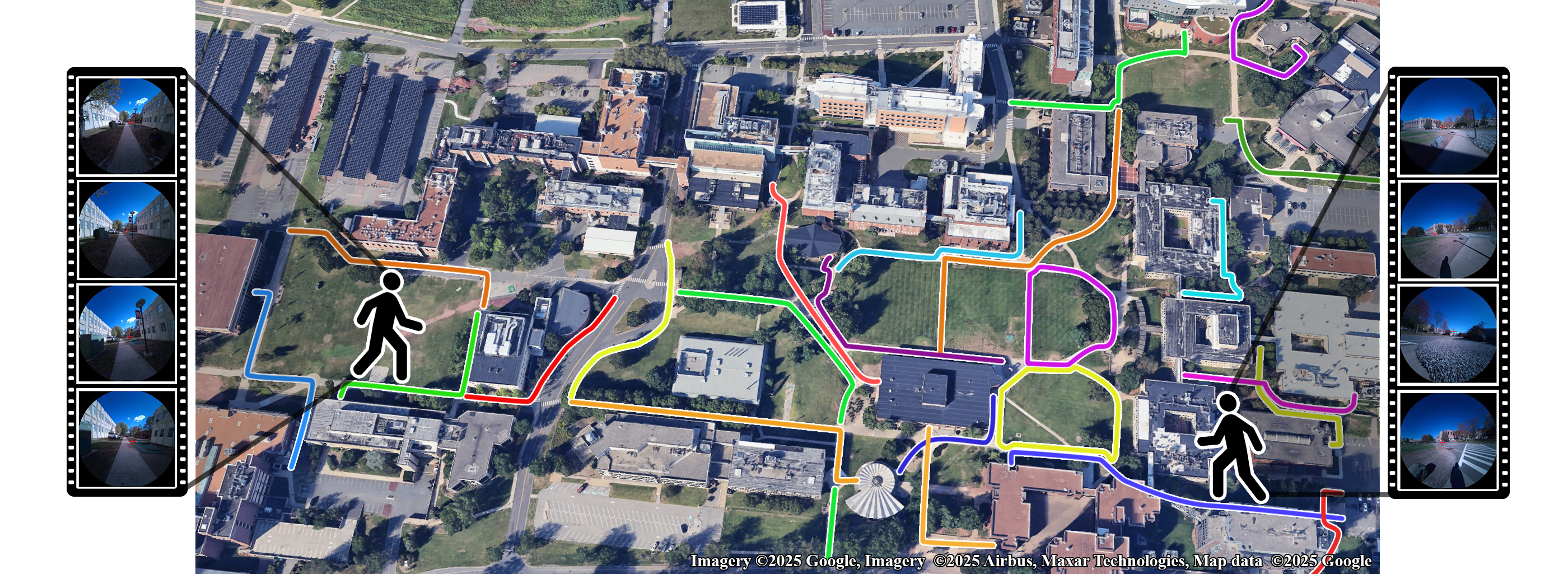}
\caption{\textbf{Dataset Paths.} A map showing a subset of the campus region, with colored paths indicating the participants' walking trajectories. During data collection, each participant follows a set of paths forwards and backwards. A sample of the captured egocentric video is shown.}
\label{fig:dataset_birdseye}
\end{figure*}

\subsection{Dataset Collection}

% project aria glasses description
\textbf{Hardware.} Our data collection hardware is the Project Aria glasses, a lightweight, non-intrusive device designed for egocentric research \cite{engel2023projectarianewtool}. We captured data from its sensor suite, including a front-facing RGB camera ($1408 \times 1408$ resolution at 30Hz) and two monochrome, global-shutter, inward-facing cameras for eye tracking. Motion and localization data were simultaneously recorded from the integrated inertial measurement units (IMUs), magnetometer, barometer, Wi-Fi and Bluetooth transceiver, and GPS. This combination of high-resolution video, gaze tracking, and motion sensors provides a rich, multimodal data stream from the human perspective. 

% our recording method 
\noindent\textbf{Recording protocol}. The dataset was collected from 82 unique subjects navigating 25 predefined paths on Busch Campus at Rutgers University. Paths ranged in length from as short as 100 meters to as long as 200 meters with an average length of ~150 meters. Data collection followed a structured protocol where each subject was instructed to walk several of these paths individually. Before starting a trajectory, participants were given a bird's-eye view map showing the start and end points, an example of which is shown in \cref{fig:dataset_birdseye}. To ensure a robust capture of real-world variability, recordings spanned different times of the day, multiple seasons, and various weather conditions, yielding a diverse set of environmental conditions.

\noindent\textbf{Data Processing.}
To prepare the dataset, we performed a multi-stage processing pipeline. First, the raw VRS files were utilized to calculate the eye tracking metrics. Per an internal review board (IRB) policy and agreements with Meta, VRS files are then modified to blur passerby in the video who did not consent to be in the dataset using EgoBlur \cite{raina2023egoblur}.  
We extracted and temporally aligned the raw VRS recordings with multimodal sensor streams. The $1408 \times 1408$ 30Hz RGB video was downscaled and saved as $224 \times 224$ pixel JPEG frames. Both the raw and downscaled video are included in our release of the dataset. Some of the sensor readings, such as the IMU, occurred at a higher frequency than the RGB camera. To create synchronized ground truth, the 30Hz eye-tracking coordinates were saved to a NumPy array with each entry corresponding to the nearest timestamped RGB frame. The high-frequency (1000Hz) IMU readings were aligned by averaging all sensor data (accelerometer, gyroscope) that occurred closest in time to each frame's timestamp and storing the result in a parallel NumPy file. This processed format facilitates efficient data loading by allowing researchers to directly index a frame and its corresponding gaze and motion data. When using data for training and inference, a set of frames constituting a video clip is sampled at a fixed interval, in a sliding window fashion. 

% IRB mention
\noindent\textbf{Privacy and Ethics.} All data collection procedures were designed to be privacy-preserving and adhered to a strict IRB protocol for human subjects research. In accordance with Project Area research guidelines, we protected both participant and pedestrian privacy by de-identifying all captured faces using the EgoBlur blurring algorithm \cite{raina2023egoblur}.

\subsection{Dataset Statistics}
The final EgoCampus dataset is a large collection of egocentric navigation, totaling approximately 32 hours of multimodal video data ($\approx$ 3.5 million frames). The data was sourced from 82 unique participants, a significant cohort size that provides a diverse set of gaze and movement behaviors. These subjects traversed a set of 25 distinct, predefined paths across a university campus. These paths were specifically chosen to cover outdoor environments on a university campus, with a rich variety of environmental contexts, lighting conditions, and interactions with static and dynamic obstacles.
A comparative analysis of the dataset compared to recent egocentric and eye gaze datasets is provided in ~\cref{tab:egodatasets}.
\newcommand{\cmark}{\textcolor{green!60!black}{\ding{51}}}
\newcommand{\xmark}{\textcolor{red!70!black}{\ding{55}}}

\begin{table}[h]
  \centering
  \footnotesize
  \setlength{\tabcolsep}{3pt}
  \begin{tabular}{@{}lccccc@{}}
    \toprule
    Dataset & Participants & \begin{tabular}{@{}c@{}}Gaze  \\ hours\end{tabular}  &  \begin{tabular}{@{}c@{}}Pedestrian \\ only\end{tabular} & IMU & GPS  \\
    \midrule

    Epic Kitchens \cite{Damen2018EPICKITCHENS}
    & 38   & 0 & \xmark & \xmark & \cmark \\

    AND \cite{pan2025lookoutrealworldhumanoidegocentric} 
    & N/A$^*$ & 4  & \cmark & \cmark & \cmark  \\

    EGTEA Gaze+ \cite{li2021eye}   
    & 32  & 28 & \xmark & \xmark & \xmark \\

    Ego4D \cite{grauman2024ego4d}        
    & 80$\dagger$  & 33  &  \xmark  & \cmark & \xmark \\
    
    GEETUP \cite{valsecchi2019introducing}  
    & 43  & 38 & \cmark & \cmark & \xmark \\

    EgoExo4D \cite{grauman2024ego}  
    & 740  & 1286 & \xmark & \cmark & \xmark \\

    EgoExoLearn \cite{huang2024egoexolearn} 
    & 136 & 120  & \xmark & \cmark & \cmark  \\

    \midrule
    
    \textbf{EgoCampus (ours)} & \textbf{82}    & \textbf{32} & \cmark & \cmark & \cmark \\
    \bottomrule
  \end{tabular}
  \caption{\textbf{A comparative analysis of EgoCampus against other key egocentric datasets.} A comparison of our dataset's attributes against prominent benchmarks in egocentric vision. The ``Pedestrian Only'' column  specifically denotes datasets that focus on outdoor pedestrian egomotion, as opposed to indoor, task-oriented scenarios (e.g., EGTEA Gaze+, Epic Kitchens). While other datasets like AND and GEETUP also capture outdoor pedestrian trajectories, EgoCampus uniquely integrates all features essential for real-world navigation research: a large-scale participant pool, extensive gaze data, and fully synchronized multimodal data, including rich inertial (IMU) sensors. \\
  $*$ AND \cite{pan2025lookoutrealworldhumanoidegocentric} does not report a number of participants. \\
  $\dagger$ As reported in \cite{Lai2024}.}
  \label{tab:egodatasets}
\end{table}

\begin{figure}[t!]
\centering
\includegraphics[width=0.95\linewidth]{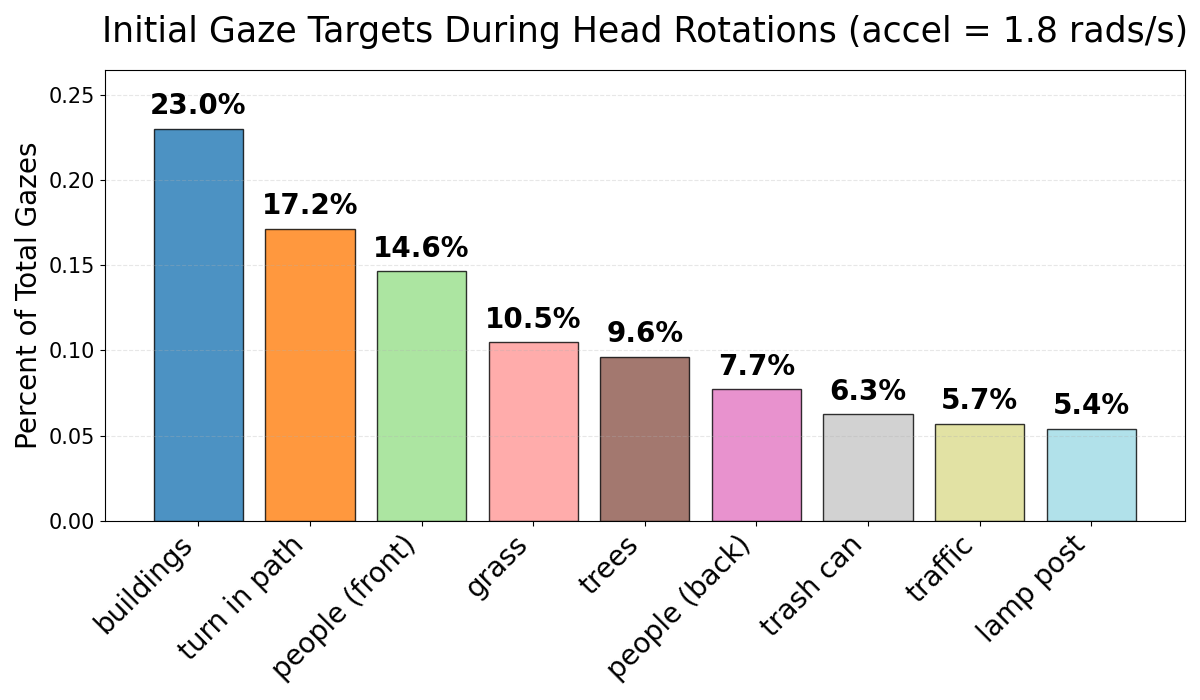}
\caption{\textbf{Attentional Targets during Head Movements.} Distribution of environmental features that capture subjects' attention during high-velocity head movements. Structural landmarks and navigational transitions constitute the majority of targets, highlighting the importance of scene understanding in gaze prediction.}
\label{fig:headmovementshistogram}
\end{figure}

\subsection{Dataset Trends}
To determine what specific environmental features capture head attention during active navigation, we conducted a manual behavioral analysis on a curated subsection of our dataset that featured rapid head rotation events. This subsection was created by utilizing a threshold of 1.8 rad/s, we isolated the exact moments where subjects shifted their gaze away from the path of travel to engage with other stimuli. This analysis was made possible by the unique multi-modal nature of the EgoCampus dataset; specifically, we leveraged the synchronized IMU (inertial measurement unit) data to filter for high-velocity rotational events. This threshold represents the lowest rotational velocity that deviates from the typical steady forward motion. As summarized in \cref{fig:headmovementshistogram}, our findings reveal that when subjects turn their heads, their attention is primarily captured by structural landmarks (buildings, trees, lamposts) and navigational cues (pedestrians, paths) were the dominant targets of interest. These high-velocity head movements constitute of approximately 12.5\% of the EgoCampus dataset, showing what people look at when they turn or what catches their attention in navigation.
\section{Methodology}\label{methodology}

\begin{figure*}[h!]
    \centering
    \includegraphics[width=0.8\linewidth]{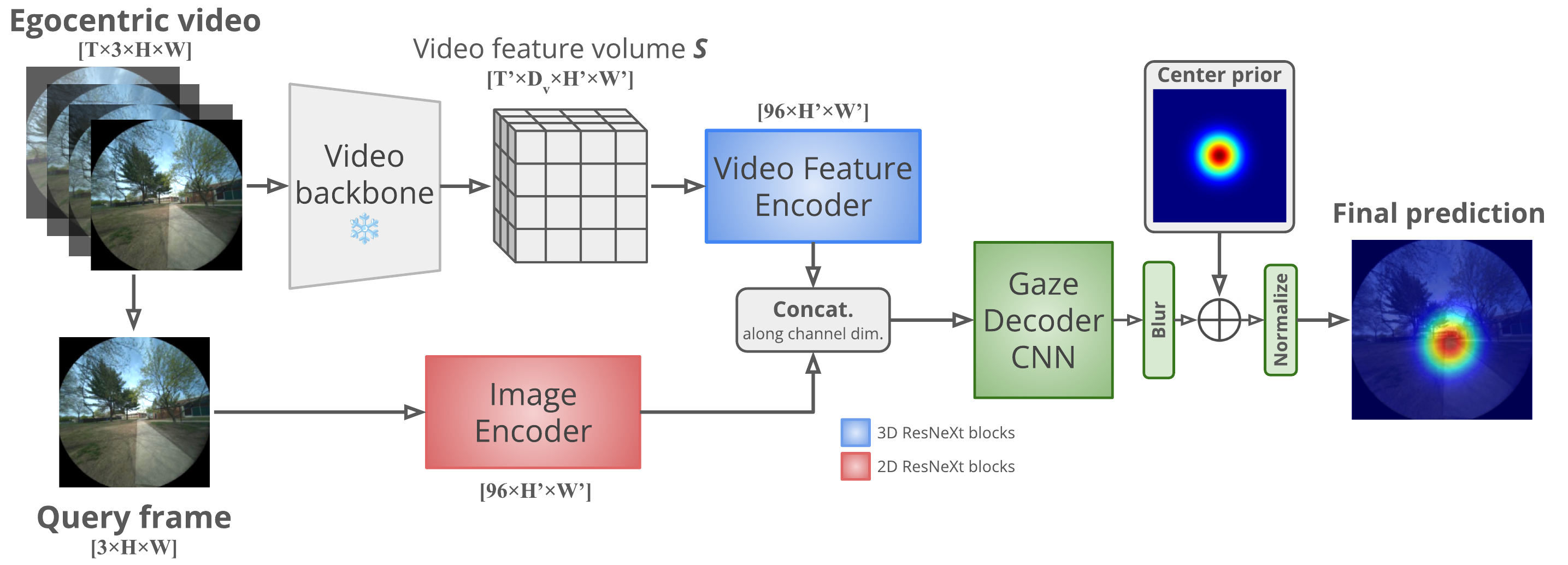}
    \caption{\textbf{An overview of the proposed spatio-temporal fusion method.} A pre-trained video feature extractor backbone is used to extract spatio-temporal features that encode information about the input frames. The spatio-temporal features are encoded and upscaled with a ResNet block. In parallel, the query frame is encoded with a seperate ResNet block. The image features and spatio-temporal features are concatenated along the feature dimension before being decoded into the final output.}
    \label{fig:stfusion}
\end{figure*}

Given an egocentric video of length $T$ and spatial dimensions $H, W$. Our goal is to predict the most likely gaze points for each frame $\{f_t \in \mathbb R^{3\times H\times W}| t<T\}$ at time $t$. The basis of our ECN approach (see ~\cref{fig:teaser}) is the assumption that information observed in the past (represented as the video) will influence a person's future gaze behavior. While videos are very high dimensional, much of the information present is redundant. This leads us to use pretrained video encoder backbones, which greatly reduce the dimensionality of our input while retaining the most useful information. For video backbones, we primarily use X3D \cite{feichtenhofer2020x3dexpandingarchitecturesefficient}. An overview of our method is presented in \cref{fig:stfusion}. Through light-weight CNN blocks, we fuse the spatio-temporal features from the video encoder backbones with learned features from a query frame. After fusion, a CNN decoder predicts the final gaze heatmap with the same spatial dimensions of the input, representing the relative likelihood of someone looking at any part of the image.

We provide the video encoder backbone with the input video and obtain spatio-temporal features $S \in \mathbb R^{N \times f_D}$ where $N= T \times \frac H p \times \frac W p$, with video encoder patch size $p$ feature dimensionality $f_D$. For predicting the gaze of frame $f_t$, the $t$-th feature in the time axis of $S$ is fed into ResNet blocks \cite{Xie2016} to obtain $S' \in \mathbb R^{H/4 \times W/4 \times 96}$. To obtain the image features, $f_t$ is fed into ResNet blocks to obtain an identically shaped feature map $I$. $S'$ and $I$ are concatenated along the channel dimension to obtain an encoded gaze map $G \in \mathbb R^{H/4 \times W/4 \times 192}$. Then, a learned CNN decoder network upscales the low-res feature map. Finally, the output is blurred, the center prior is added, and a normalization step produces the predicted gaze heatmap.

Models were implemented and trained on a single NVIDIA RTX 3090 GPU in the PyTorch framework. The ADAM optimizer \cite{adam_optimizer} was used with an MSE loss and a learning rate of 0.002. Our final model trained for 10 epochs in approximately 8 hours.
\section{Experiments}
We first review the evaluation metrics we use for gaze prediction in \cref{sec:metrics}. We also compare EgoCampusNet (ECN) to existing models from the domains of image saliency, video saliency, and eye gaze prediction in \cref{sec:comparisons}. Finally, we showcase qualitative results and discuss the observed behaviors of the model in \cref{sec:qualitative}. We present the results of our model trained on our data, compared to the performance of existing pretrained models, which we do not train on our data. To demonstrate the need for further research in this area, we show that existing SOTA models do not generalize well enough to have strong performance on our dataset without fine tuning.

For our experiments, we sample the EgoCampus dataset 16 frames at a time, uniformly sampled from a window of 64 frames ($\approx$ 2.1 seconds). Models that utilize temporal information are provided with the full video clip in order to predict gaze in the last frame, while image-based models are only provided with the last frame for prediction. The test/train split follows an approximate 70/30 ratio, where a randomly selected set of 16 paths are resevered for training, and the remaining 9 paths are used for testing. Exact splits will be provided in supplementary material.

\subsection{Evaluation Metrics}\label{sec:metrics}
We distinguish between saliency prediction, which models ``free-viewing'' fixation density, and eye gaze prediction, which captures the goal-directed navigation behavior of EgoCampus.To assess performance, we utilize \textbf{location-based} metrics, such as AUC-Judd, which use binary fixation maps to evaluate the identification of specific salient points, and \textbf{distribution-based} metrics, including CC, SIM, and KLD, which assess the statistical similarity between predicted and ground truth distributions. Larger values indicate better alignment for all metrics except KLD, where lower values reflect higher accuracy.

\vspace{-0.1cm}

\paragraph{Area Under Curve (AUC-Judd):} AUC-Judd measures how well the predicted saliency map distinguishes between fixated and non-fixated locations using a binary ground truth fixation map. It computes the area under the Receiver Operating Characteristic (ROC) curve. This curve is generated by plotting the True Positive Rate (TPR) (the hit rate, or the percentage of fixations correctly identified as positive) against the False Positive Rate (FPR) while varying a threshold on the predicted saliency values. AUC-Judd is independent of thresholds, making it a robust metric, but it can become saturated in the eye gaze setting. Higher values indicate stronger model performance.

\begin{equation}
    AUC_{\text{Judd}} = \sum_{i=1}^{n-1} \frac{(\text{TPR}_i + \text{TPR}_{i+1})}{2} \times (\text{FPR}_{i+1} - \text{FPR}_i)
    \label{eq:auc_trapz}
\end{equation}

\paragraph{Correlation Coefficient (CC):}CC captures the linear correlation between the pixel values of the predicted saliency map ($P$) and the ground truth distribution ($Q$). Unlike metrics that use a binary ground truth, CC requires a continuous ground truth map ($Q$), which is typically created by applying a Gaussian blur to the binary fixation points. This creates a ground truth density map. The metric then treats $P$ and $Q$ as two long vectors of pixel intensities.

\begin{equation}
\text{CC} = \frac{\text{cov}(P, Q)}{\sigma(P) \times \sigma(Q)}
\label{eq:cc}
\end{equation}

As shown in the formula, the score is calculated by taking the covariance of $P$ and $Q$ (how their pixel values vary together) and normalizing it by the product of their individual standard deviations ($\sigma$). This assesses whether the spatial patterns co-vary. A high positive value (near 1) implies strong similarity, and a score near 0 indicates no linear relationship.

\paragraph{Kullback-Leibler Divergence (KLD):}

\begin{equation}
    \text{KLD} (P \parallel  Q) = \sum_{i} Q(i)  \log \left( \frac{Q(i)}{P(i) + \epsilon} \right) ,
    \label{eq:kld}
\end{equation}

KLD measures the dissimilarity between the predicted ($P$) and ground truth ($Q$) saliency distributions. To create these distributions, both maps are first converted into probability mass functions (PMFs) where all pixel values sum to 1. $\epsilon$ is a small constant added for numerical stability, preventing division by zero where a ground truth fixation $Q(i)$ exists but the prediction $P(i)$ is zero. Lower KLD values reflect more accurate predictions.

\begin{figure*}[t!]
  \centering
  \setlength\tabcolsep{2pt}
  \begin{tabular}{*{9}{c}}
  
    %=== Row 1 ===
    \includegraphics[width=0.11\textwidth]{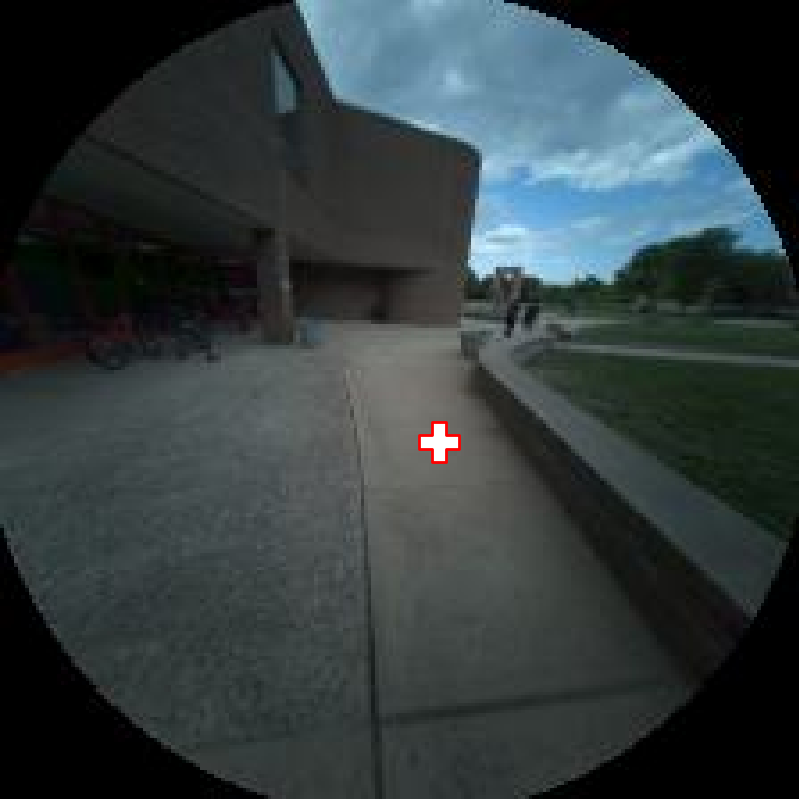} &
    \includegraphics[width=0.11\textwidth]{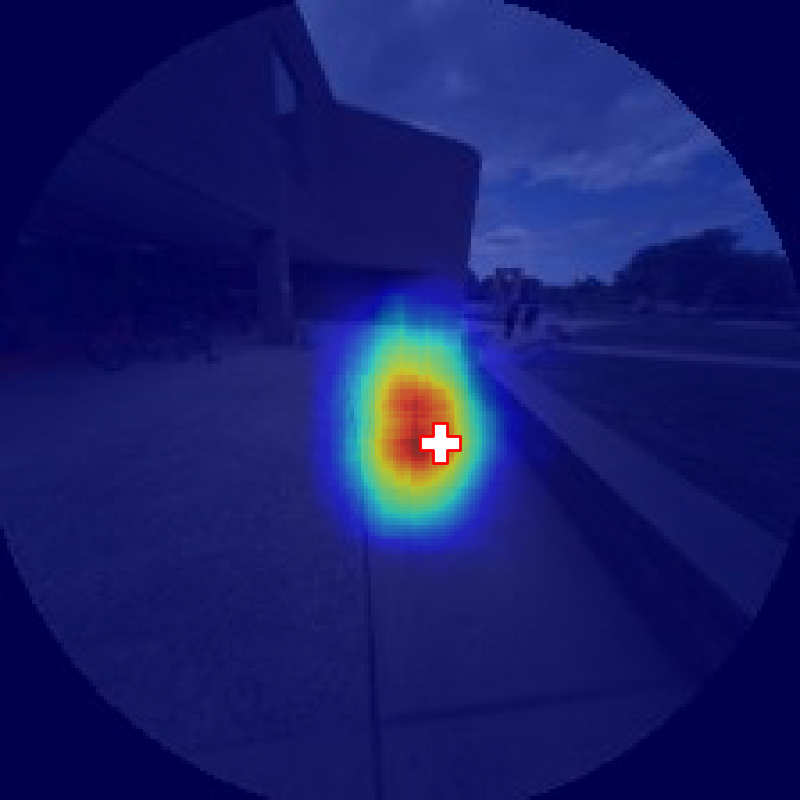} &
    \includegraphics[width=0.11\textwidth]{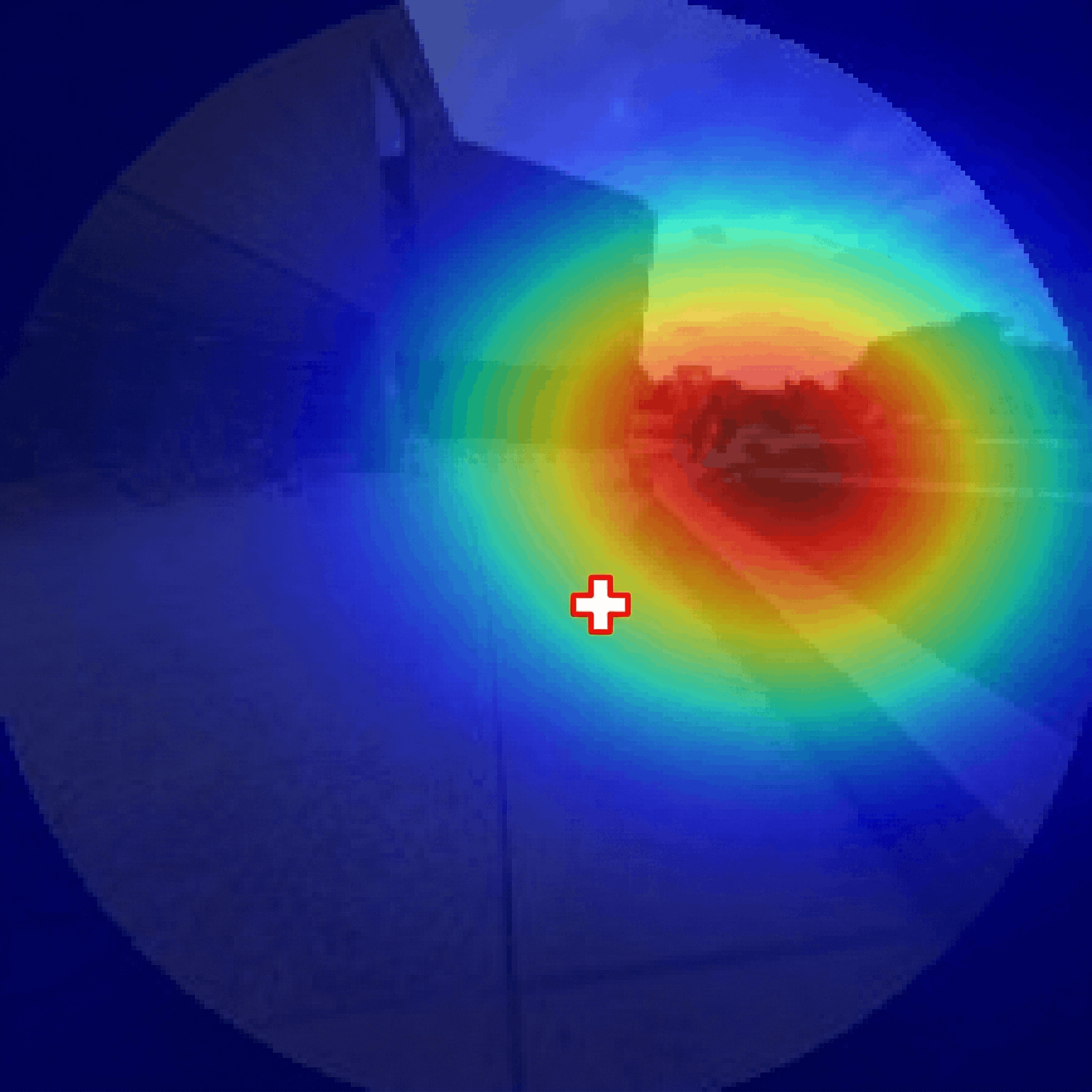} &
    \includegraphics[width=0.11\textwidth]{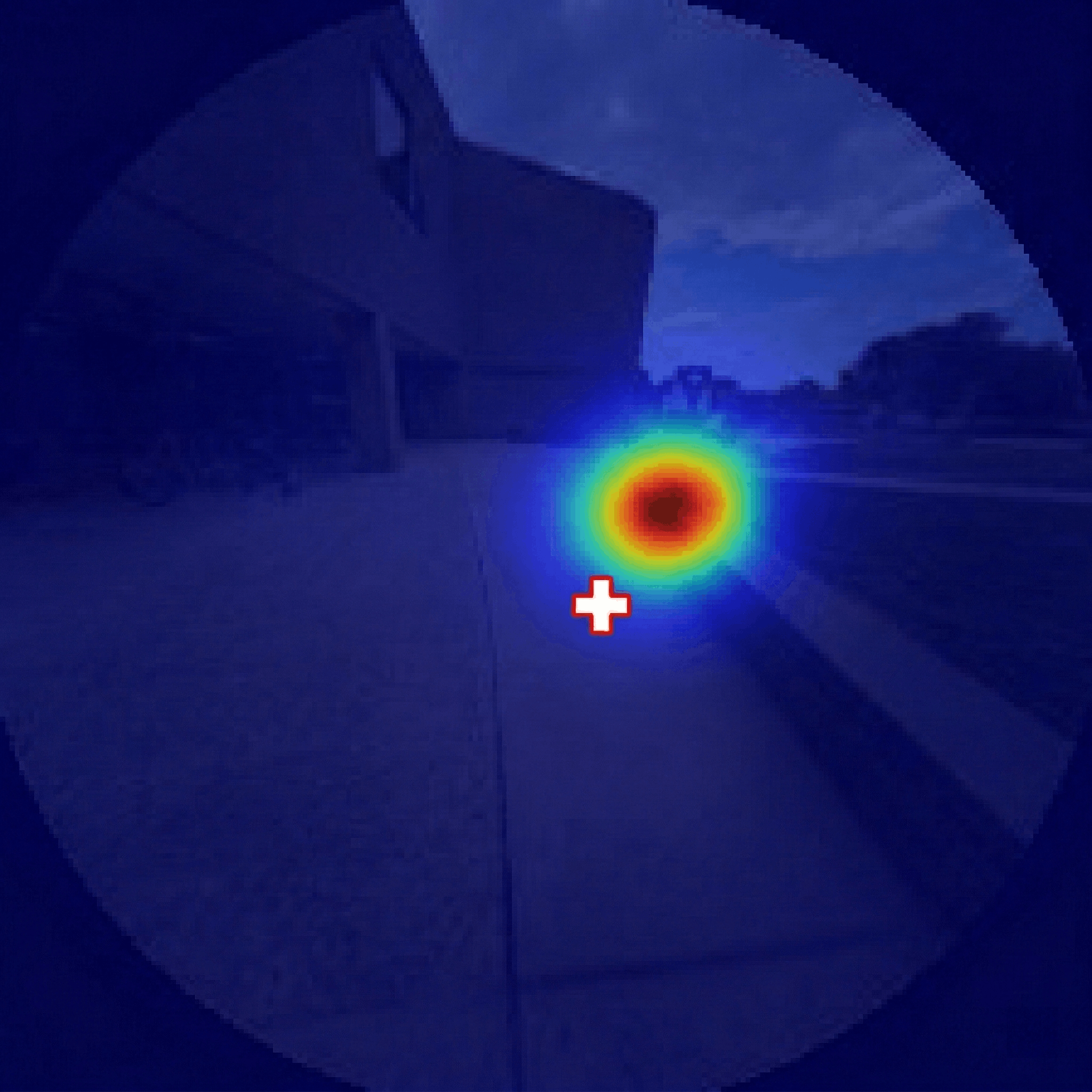} &
    \includegraphics[width=0.11\textwidth]{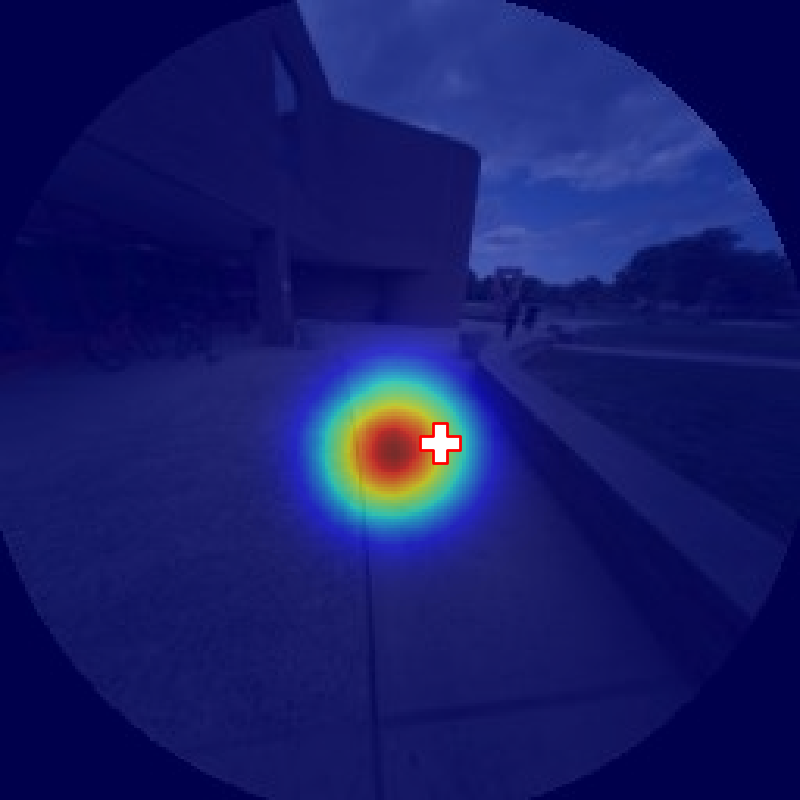} &
    \includegraphics[width=0.11\textwidth]{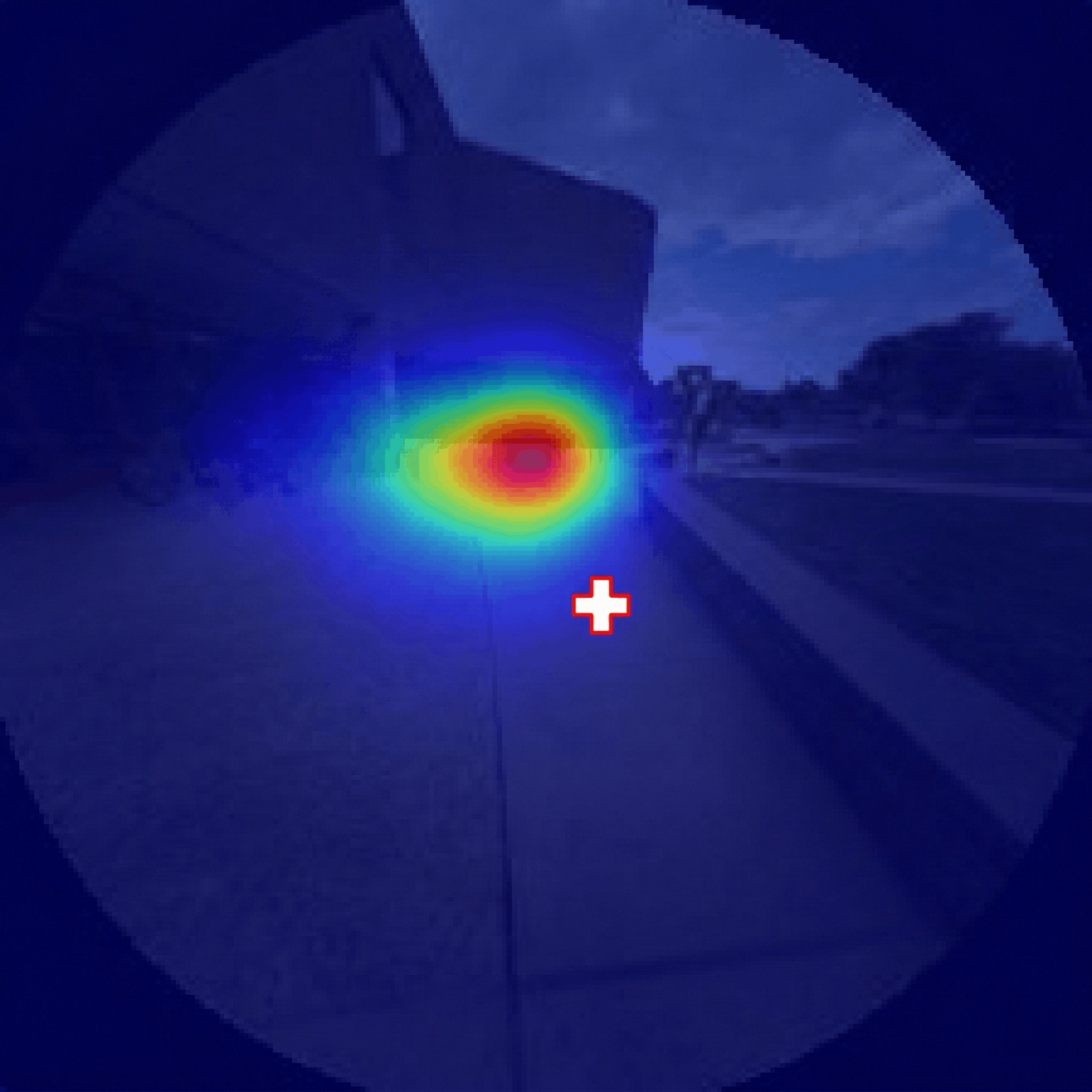}&
    \includegraphics[width=0.11\textwidth]{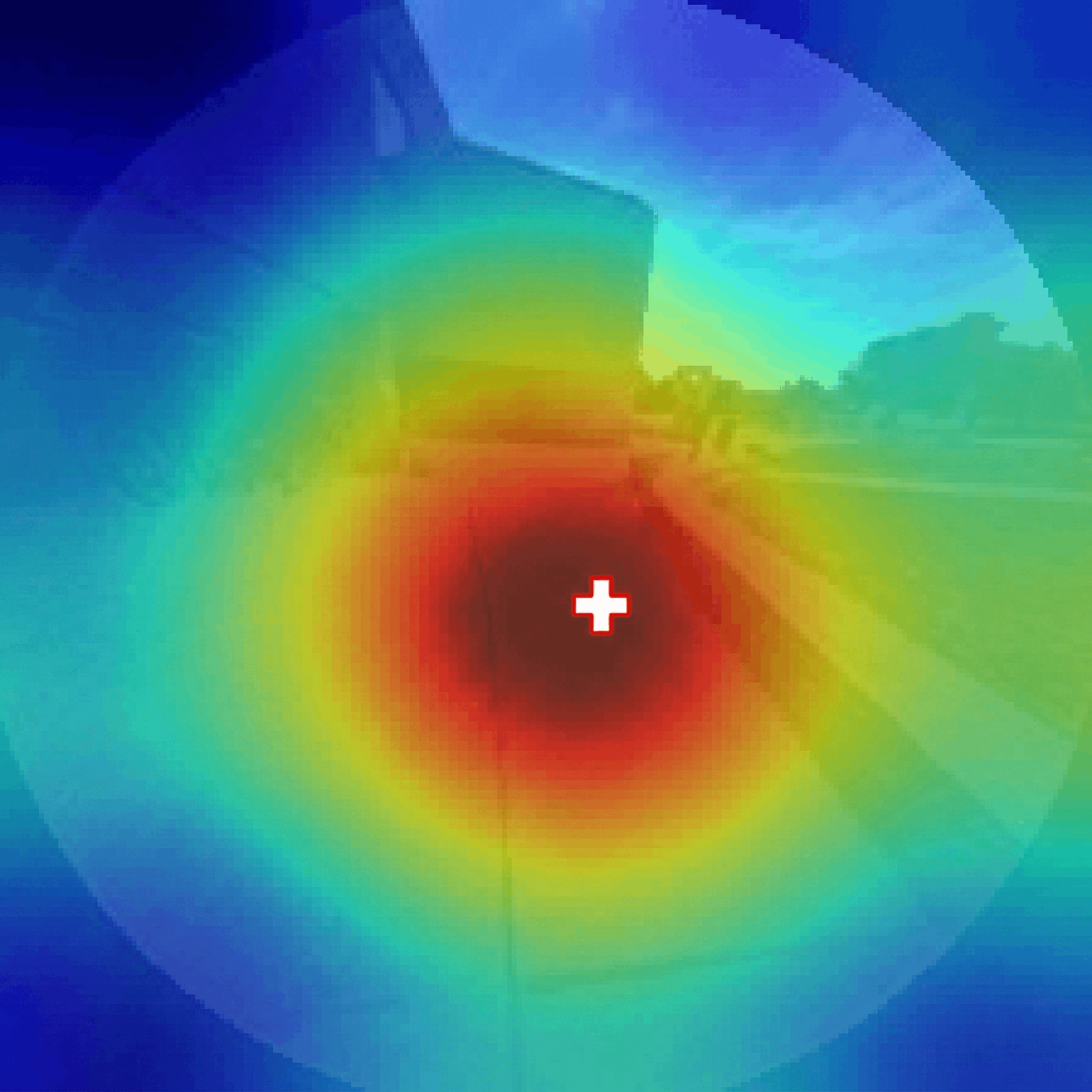} &
    \includegraphics[width=0.11\textwidth]{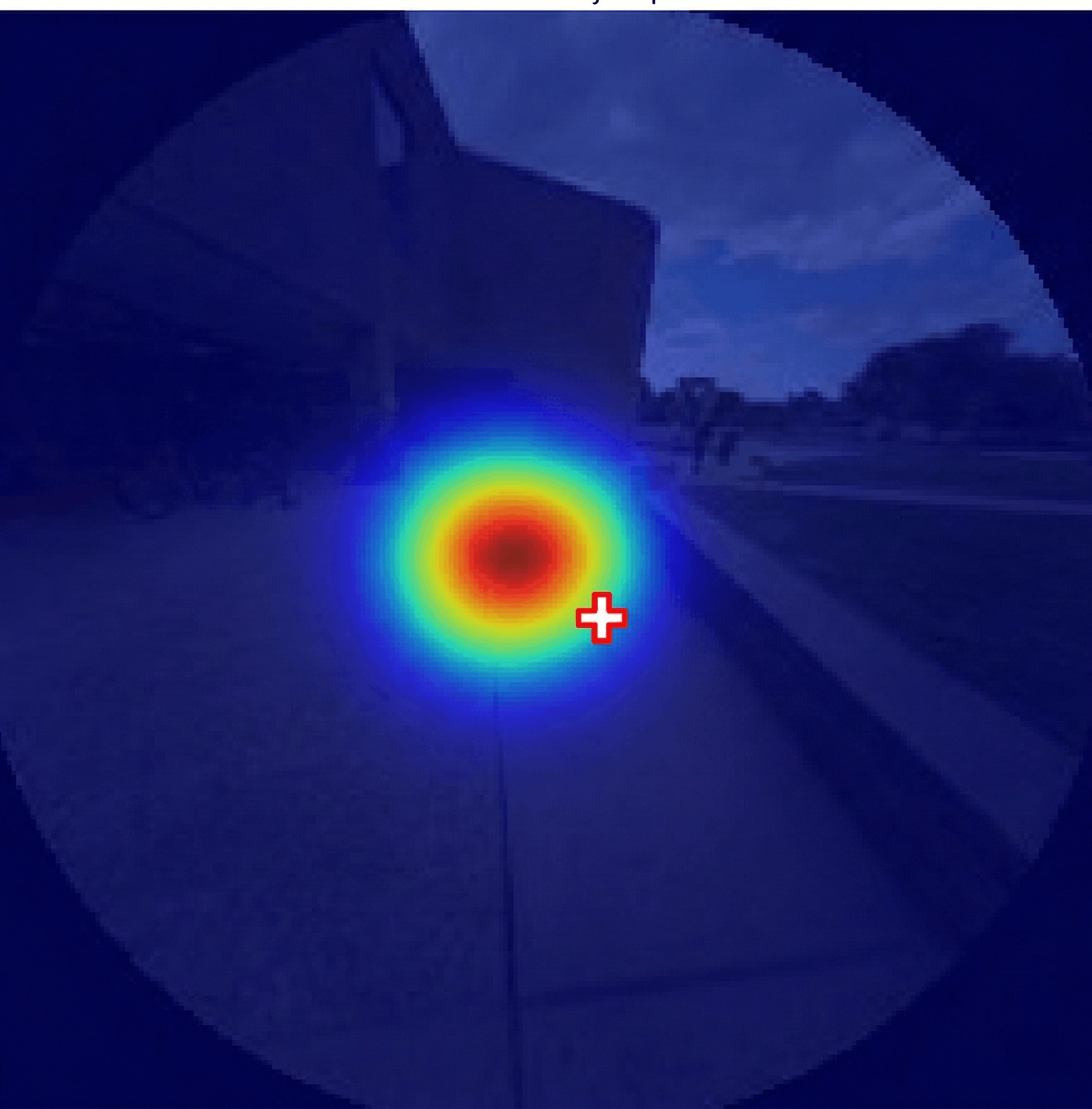}
    \\
    
    %=== Row 2 ===
    \includegraphics[width=0.11\textwidth]{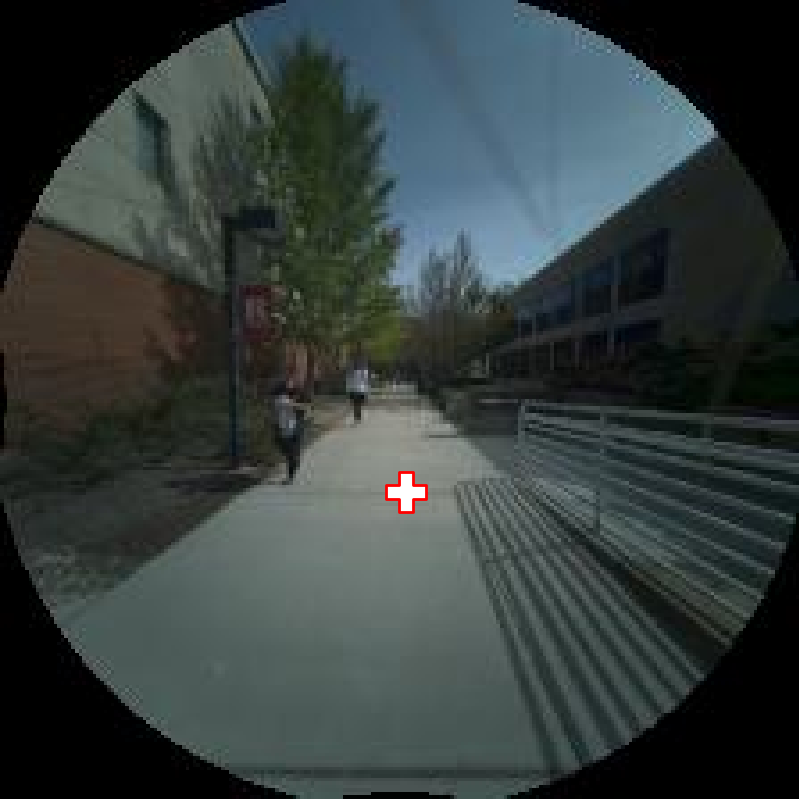} &
    \includegraphics[width=0.11\textwidth]{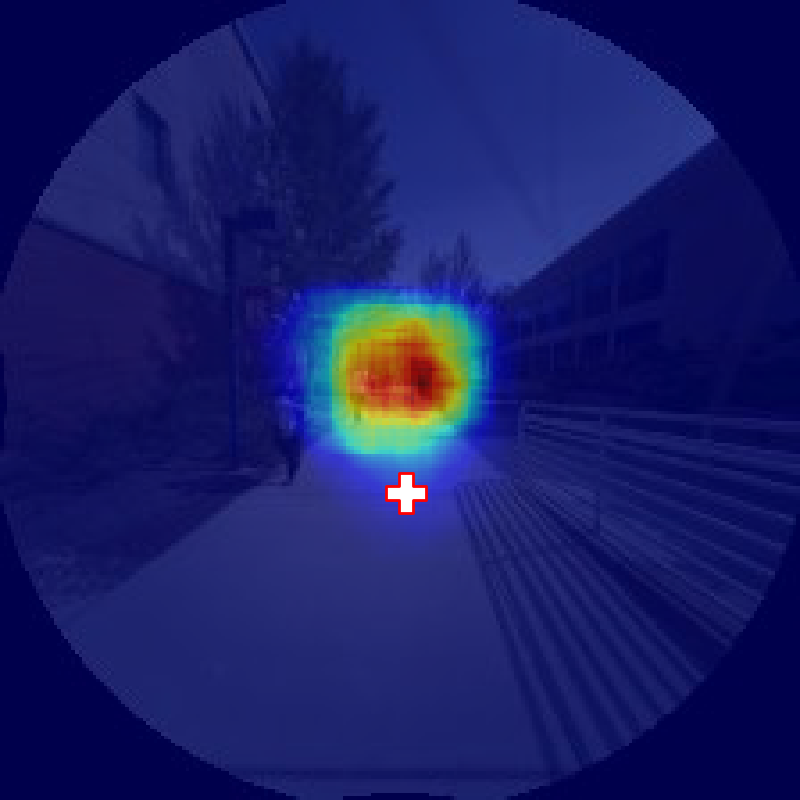} &
    \includegraphics[width=0.11\textwidth]{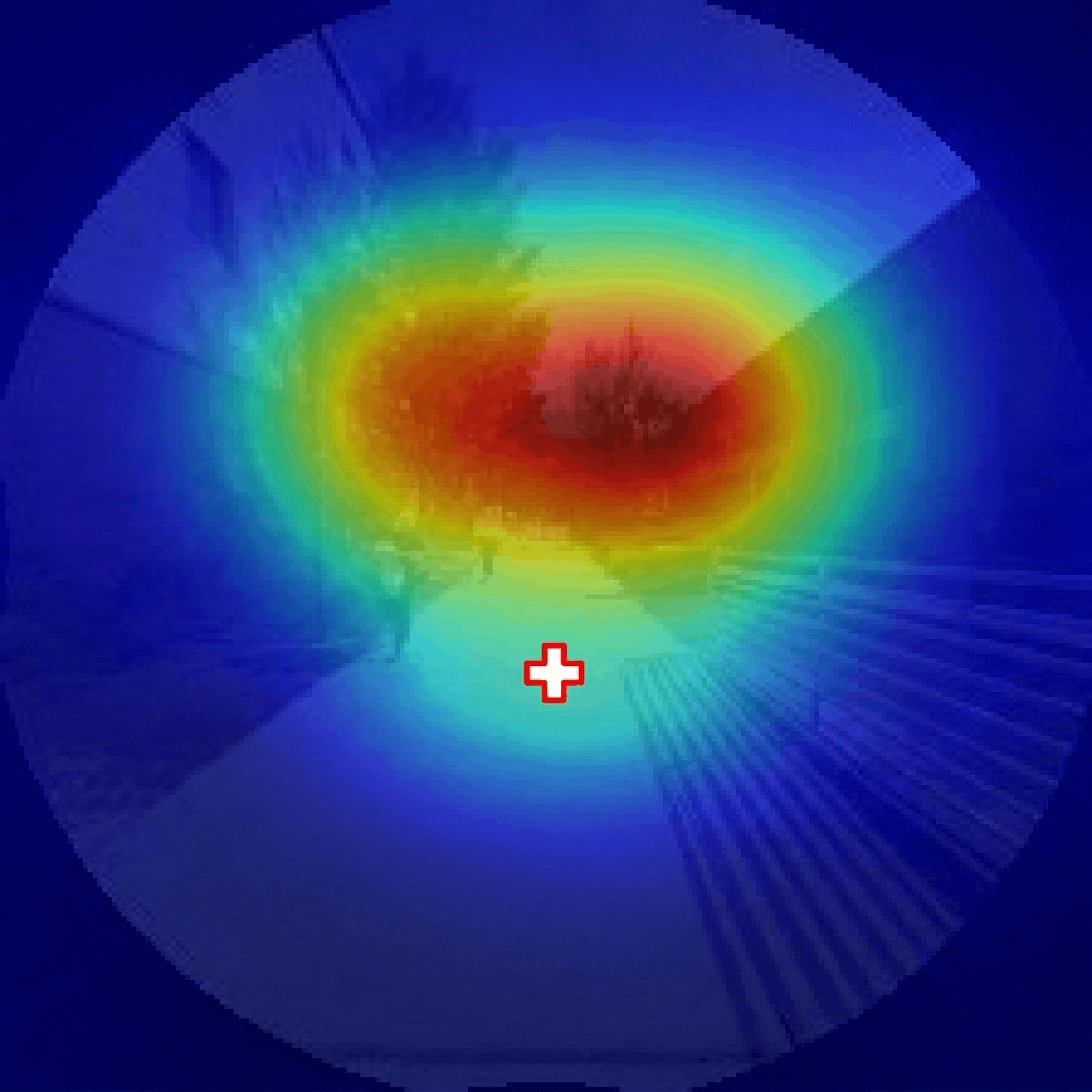} &
    \includegraphics[width=0.11\textwidth]{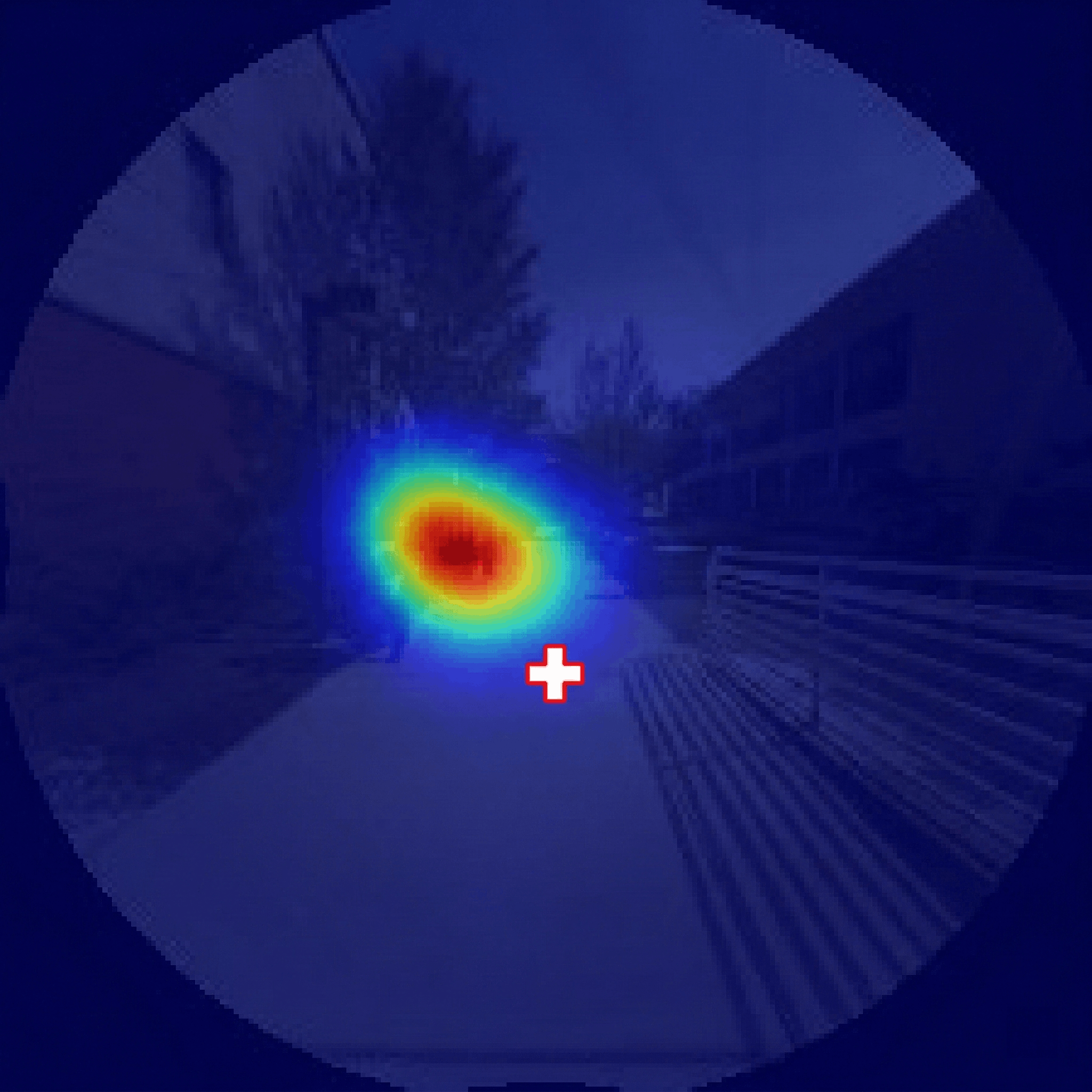} &
    \includegraphics[width=0.11\textwidth]{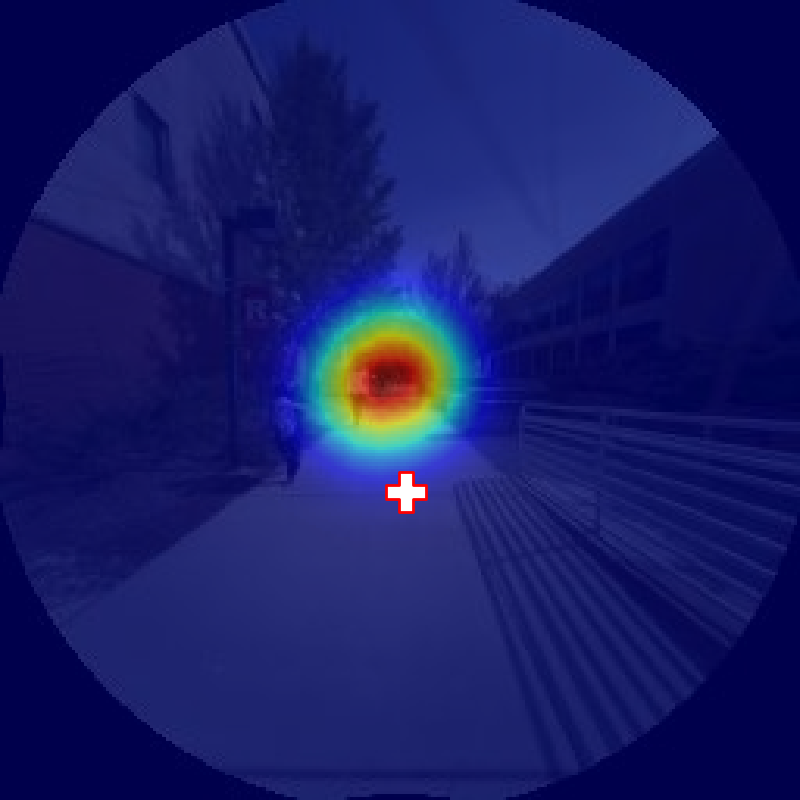} &
    \includegraphics[width=0.11\textwidth]{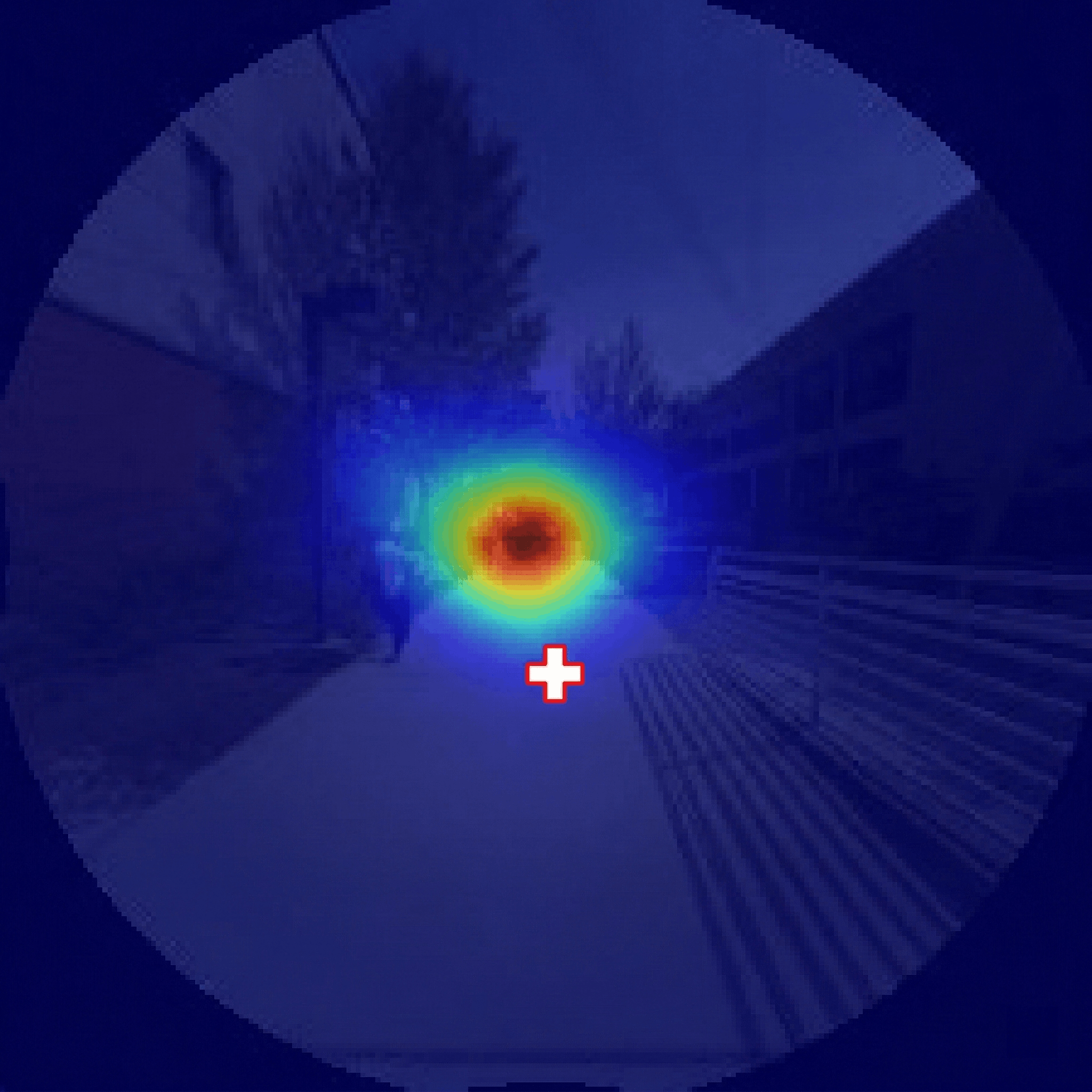}&
    \includegraphics[width=0.11\textwidth]{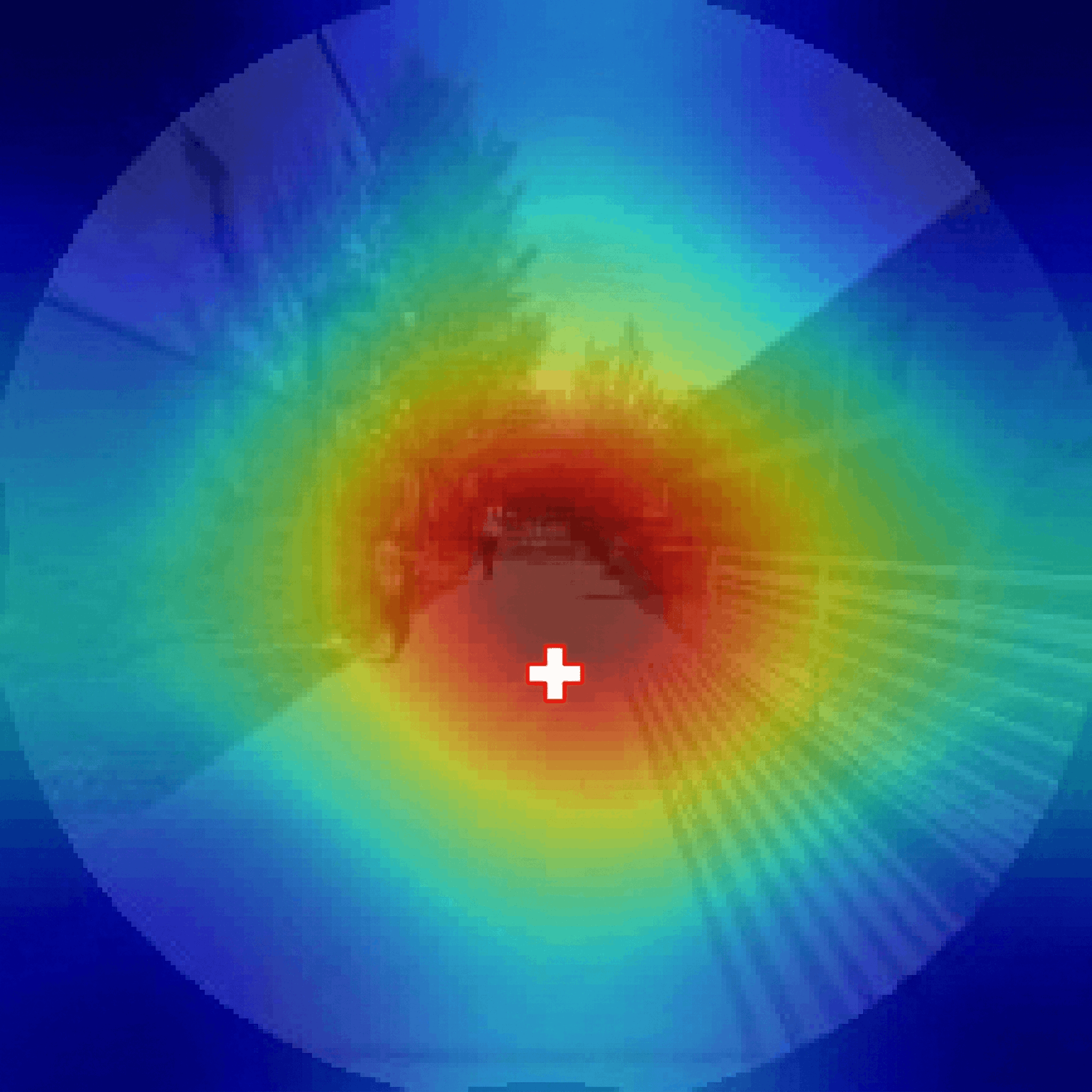} &
    \includegraphics[width=0.11\textwidth]{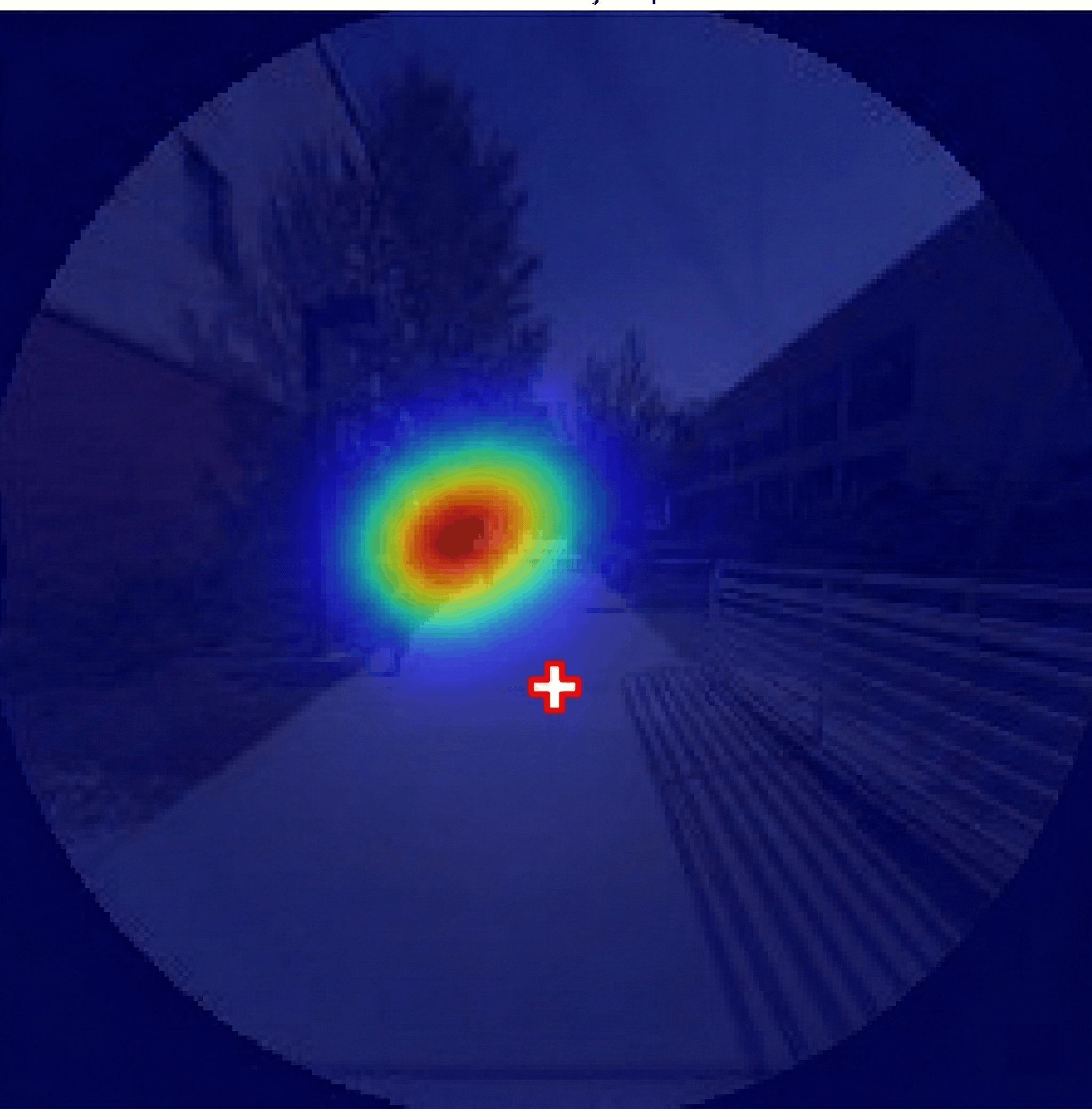}
    \\
    
    %=== Row 3 ===
    \includegraphics[width=0.11\textwidth]{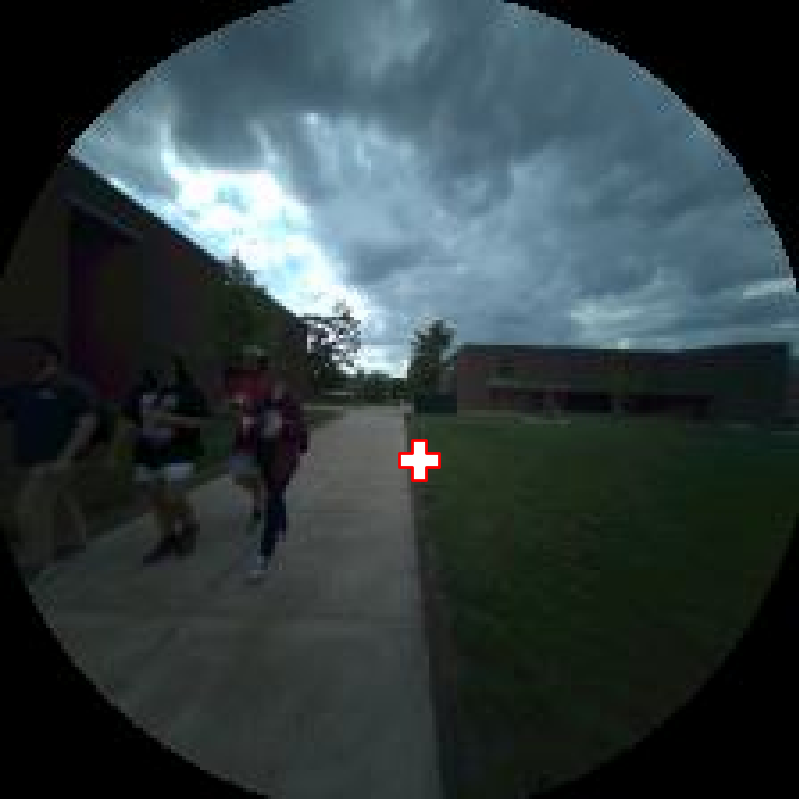} &
    \includegraphics[width=0.11\textwidth]{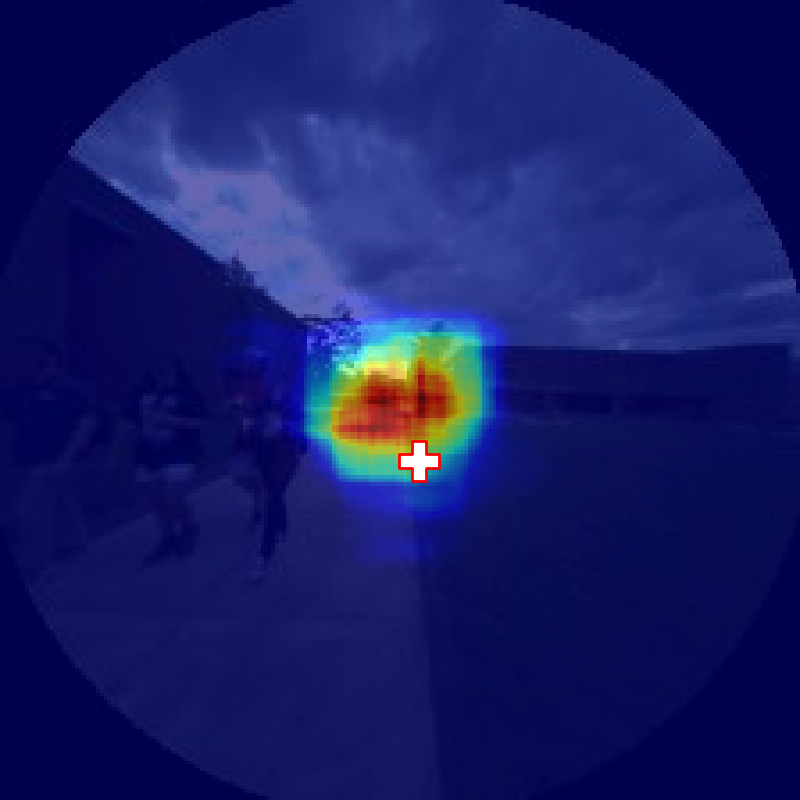} &
  \includegraphics[width=0.11\textwidth]{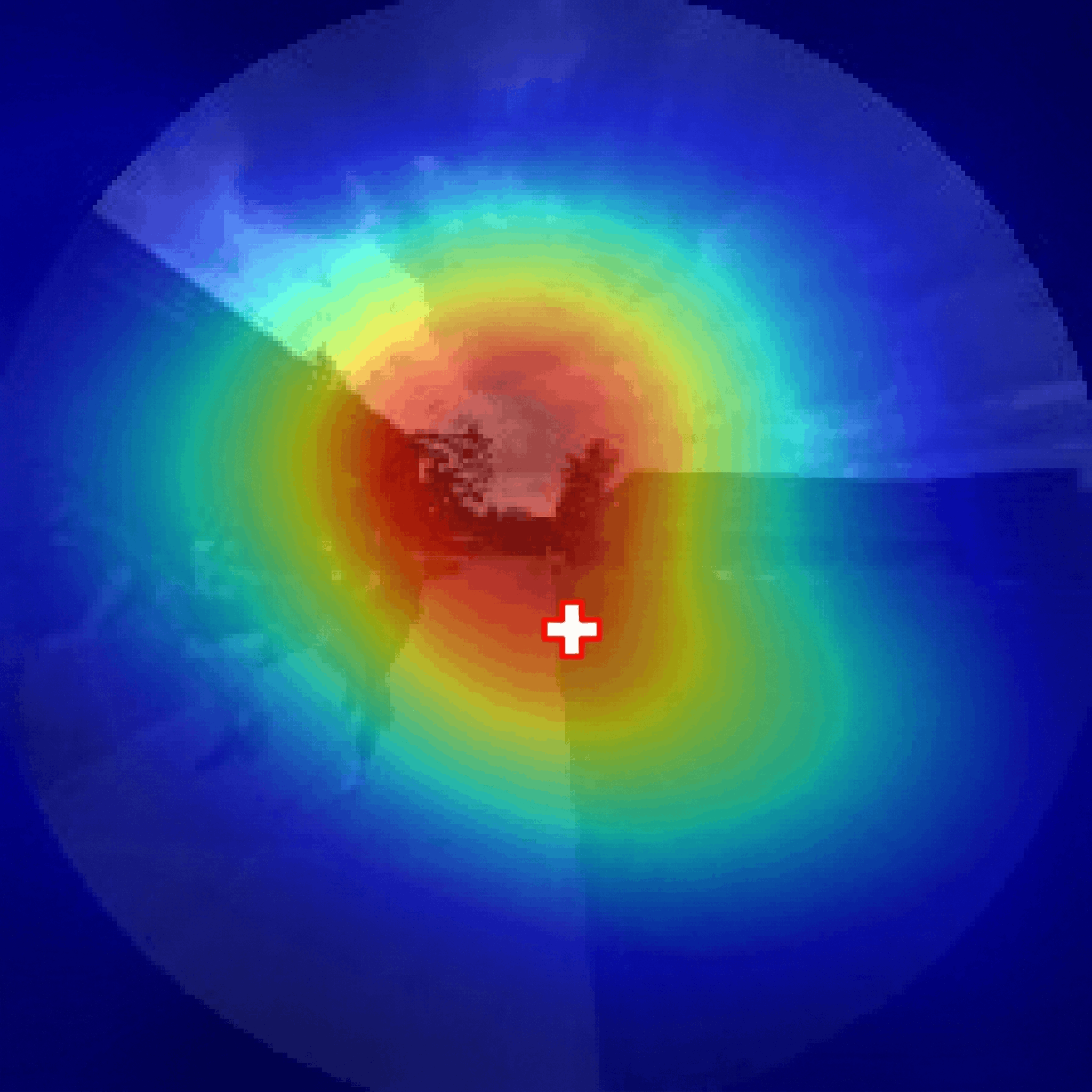} &
    \includegraphics[width=0.11\textwidth]{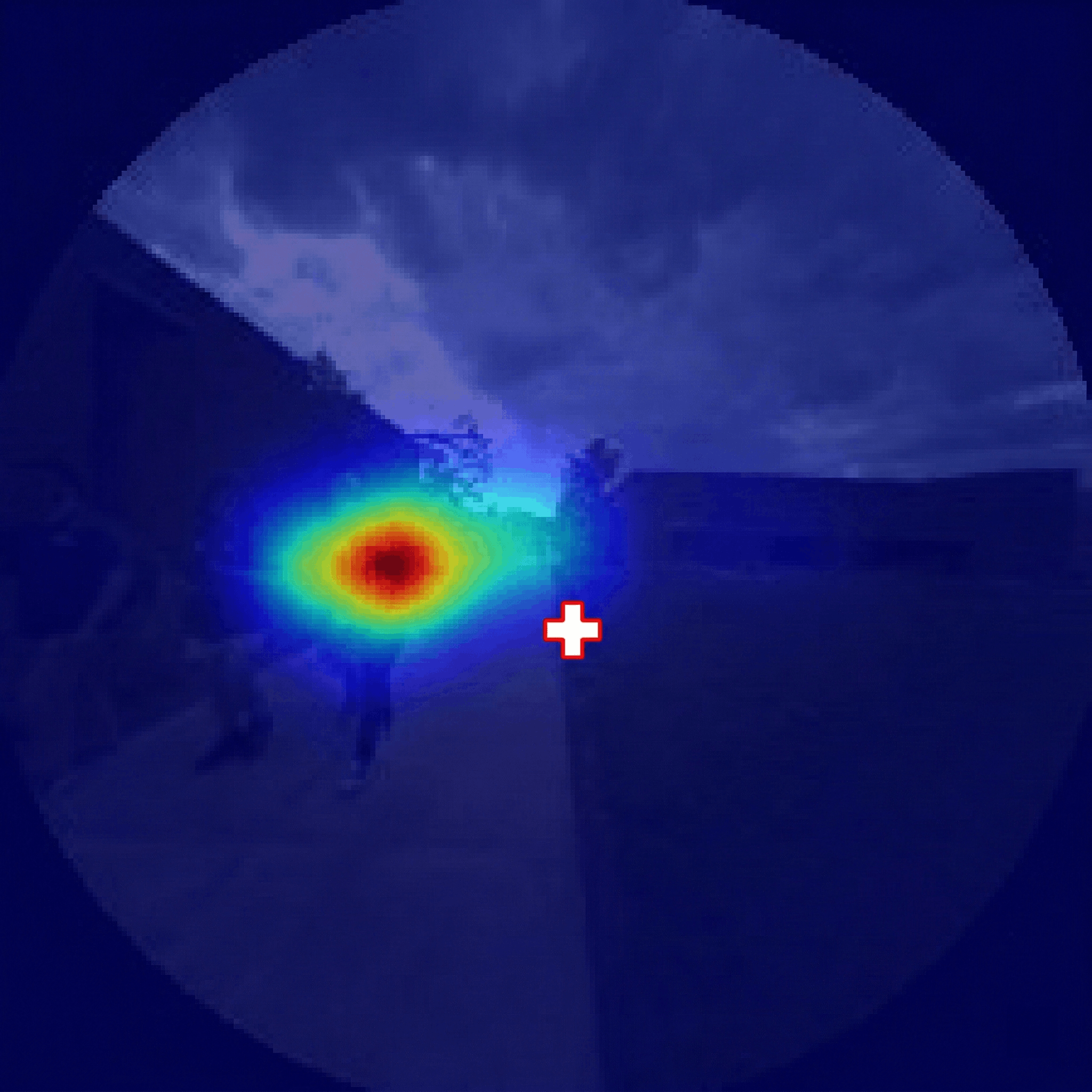} &
    \includegraphics[width=0.11\textwidth]{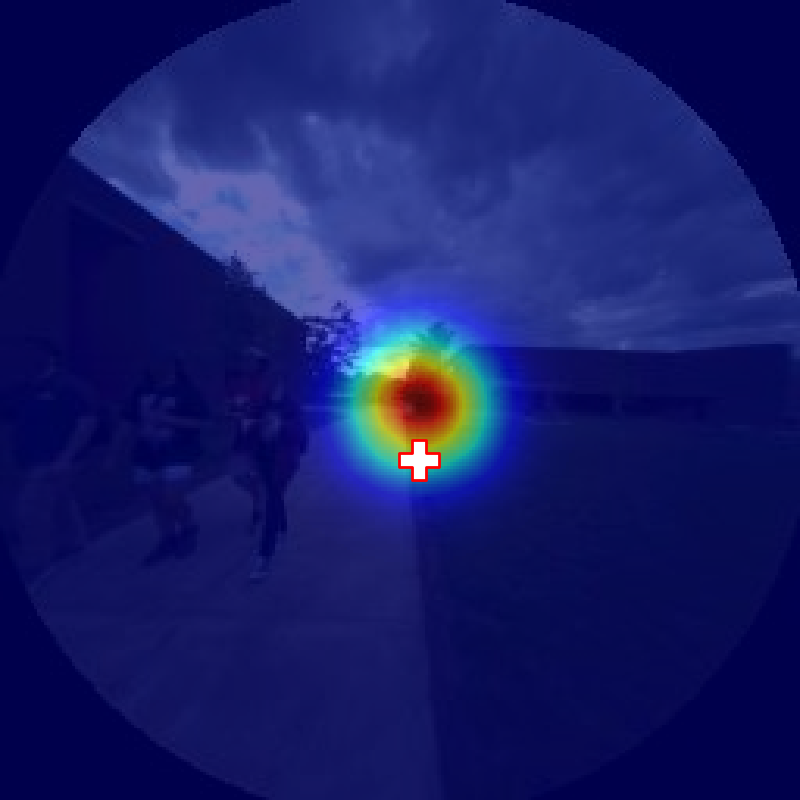} &
    \includegraphics[width=0.11\textwidth]{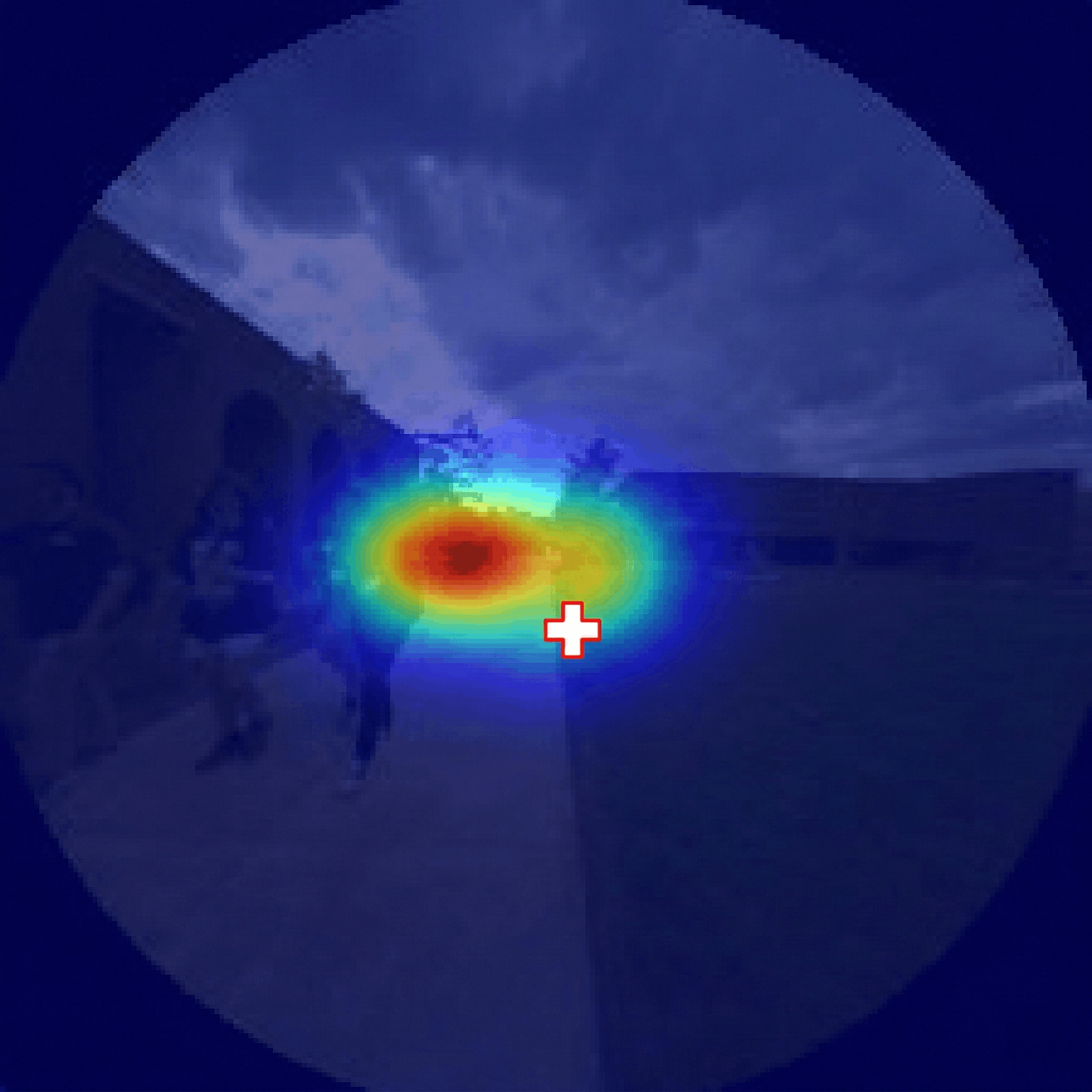}&
    \includegraphics[width=0.11\textwidth]{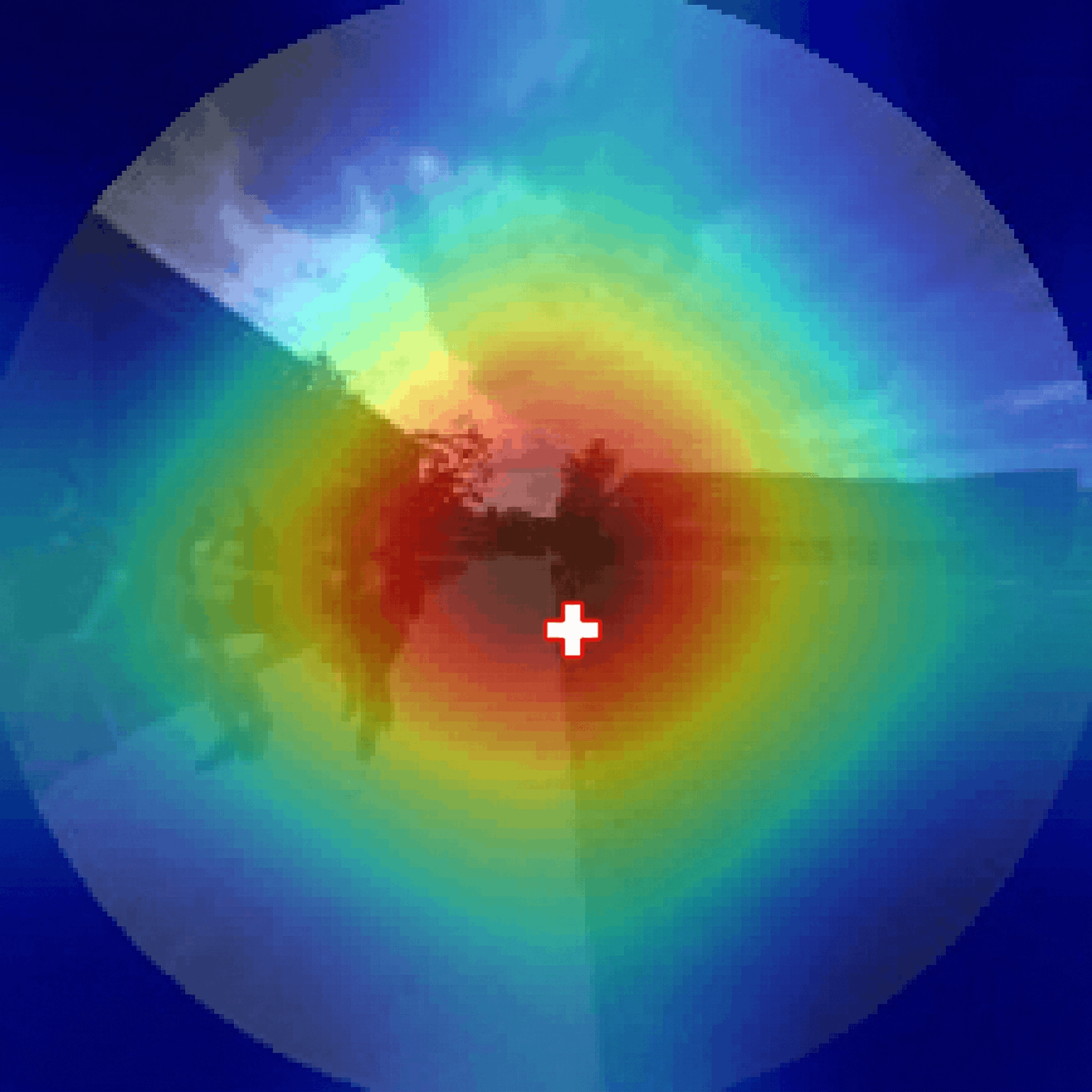} &
    \includegraphics[width=0.11\textwidth]{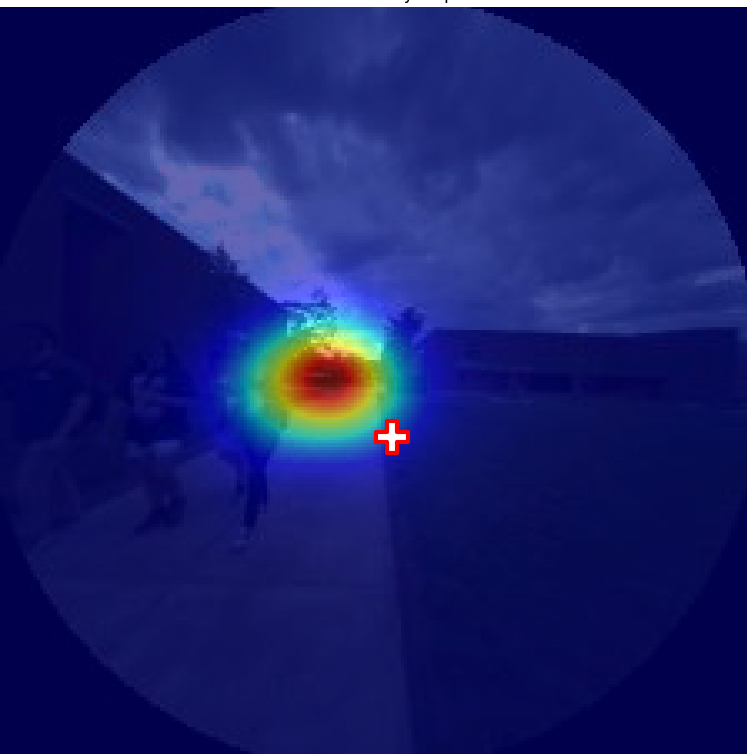}
    \\

    %=== Row 4 ===
    \includegraphics[width=0.11\textwidth]{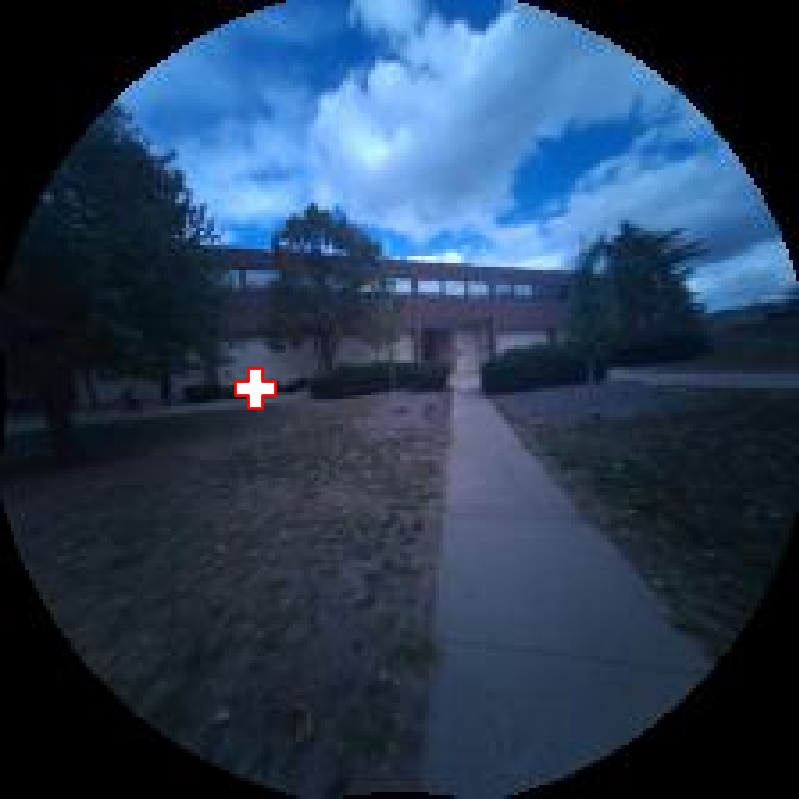} &
    \includegraphics[width=0.11\textwidth]{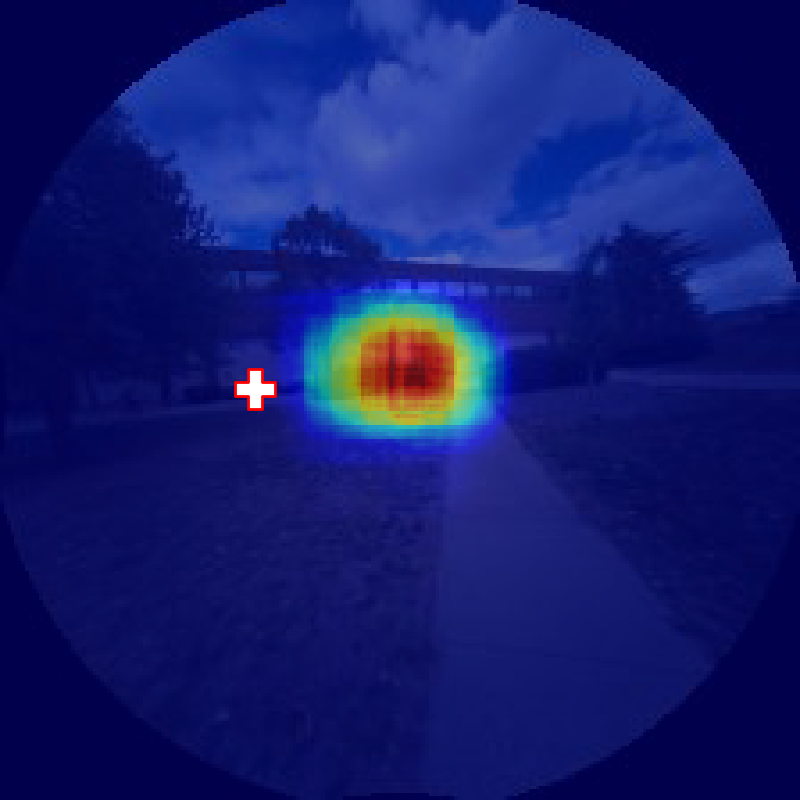} &
    \includegraphics[width=0.11\textwidth]{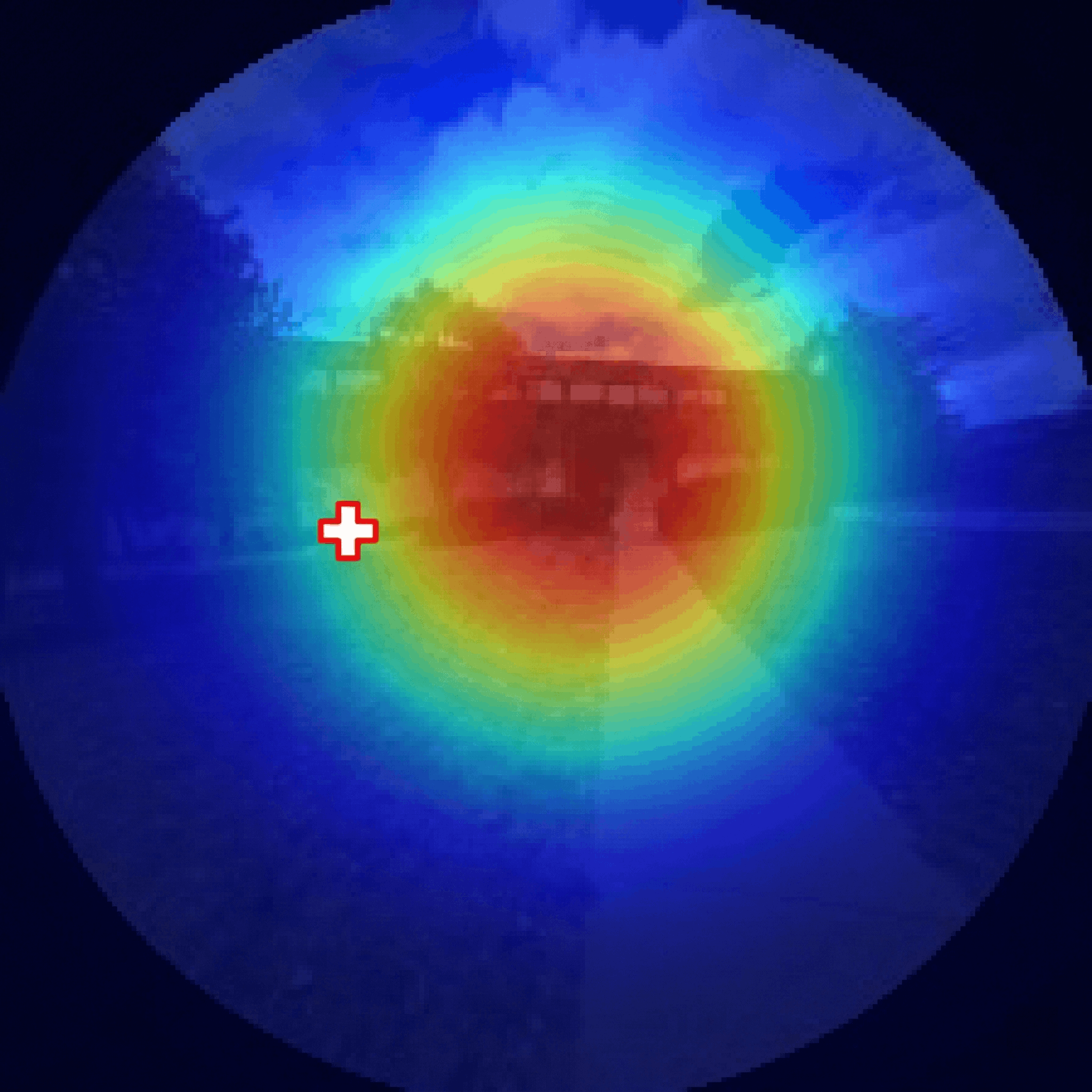} &
    \includegraphics[width=0.11\textwidth]{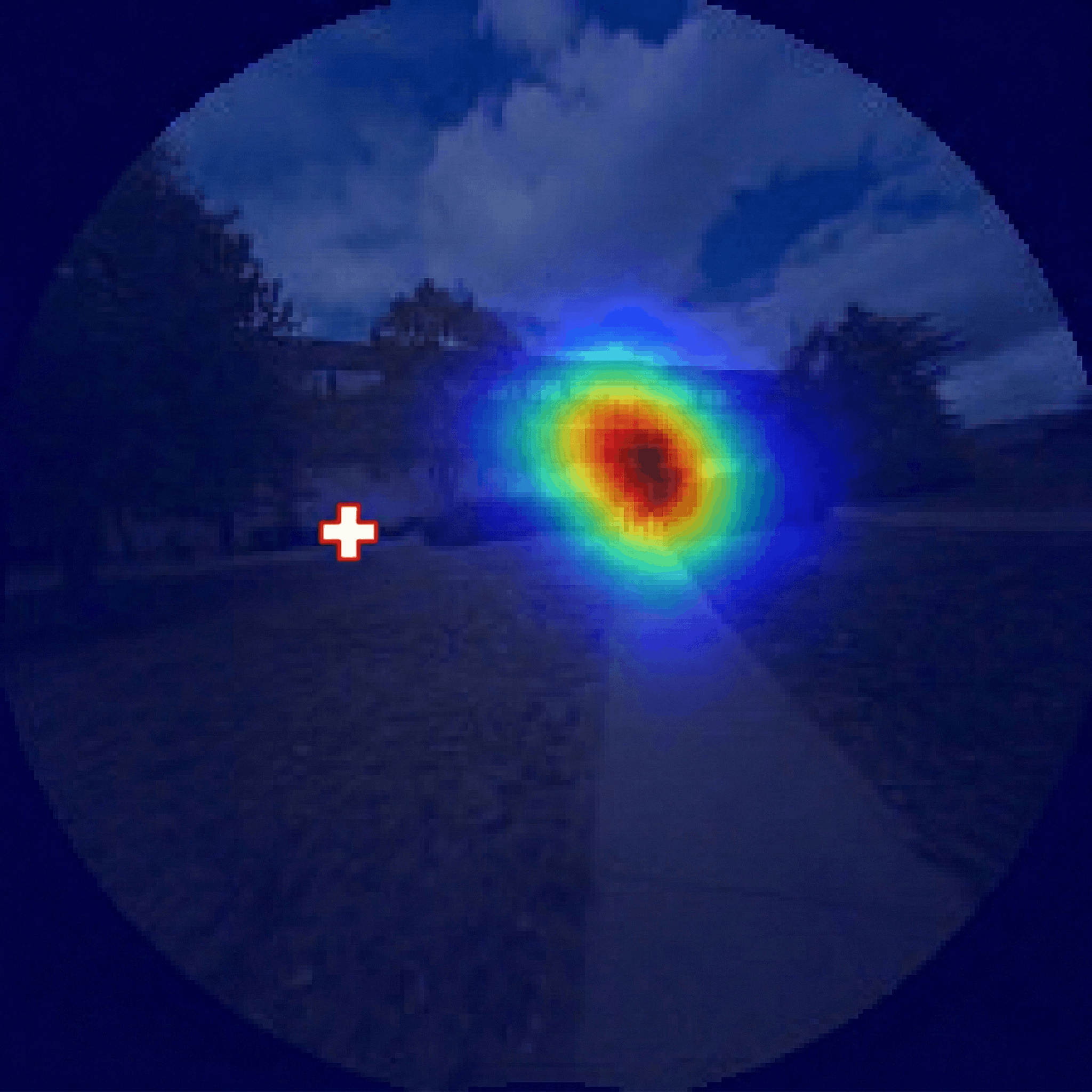} &
    \includegraphics[width=0.11\textwidth]{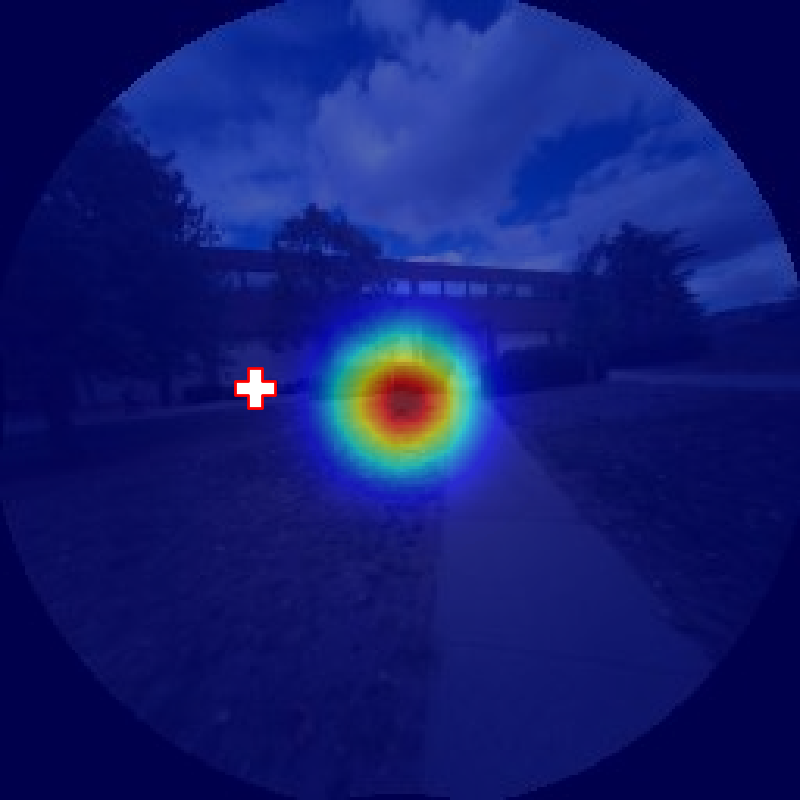} &
    \includegraphics[width=0.11\textwidth]{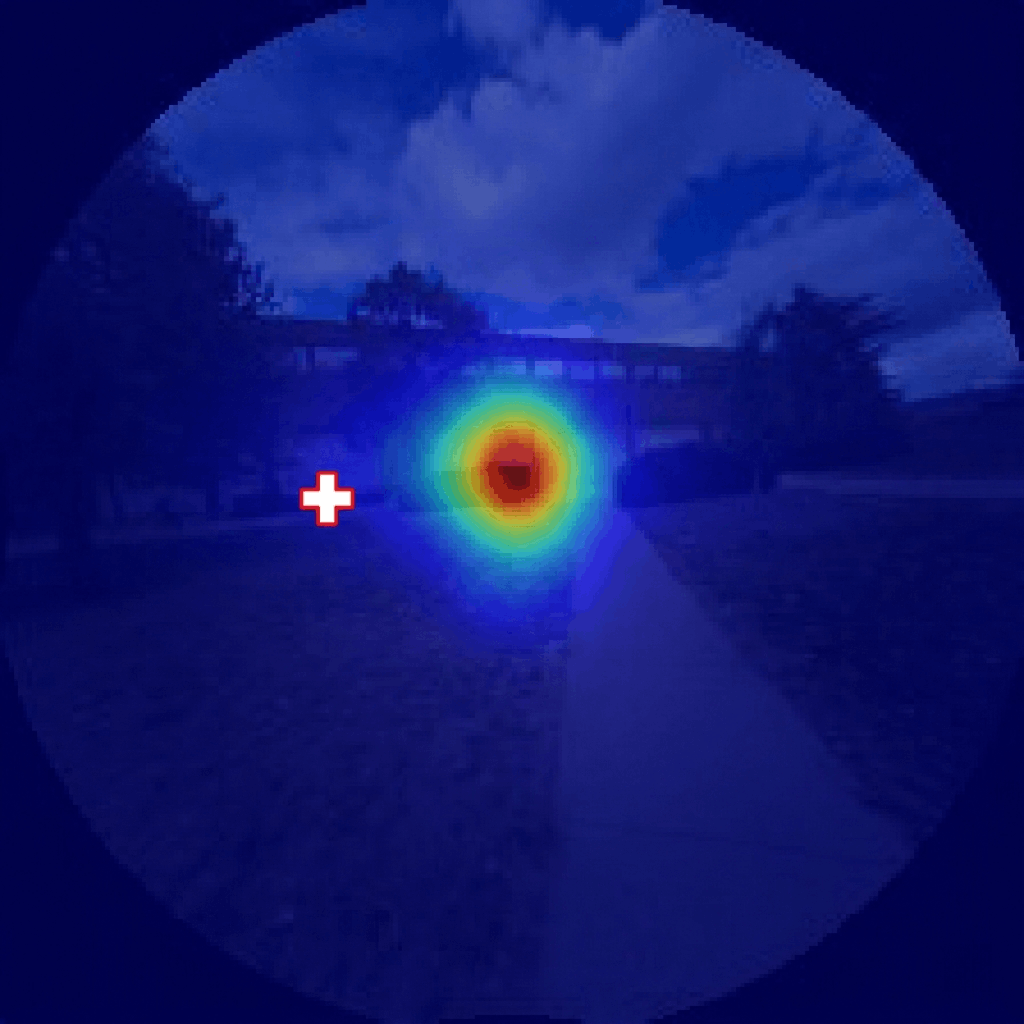}&
    \includegraphics[width=0.11\textwidth]{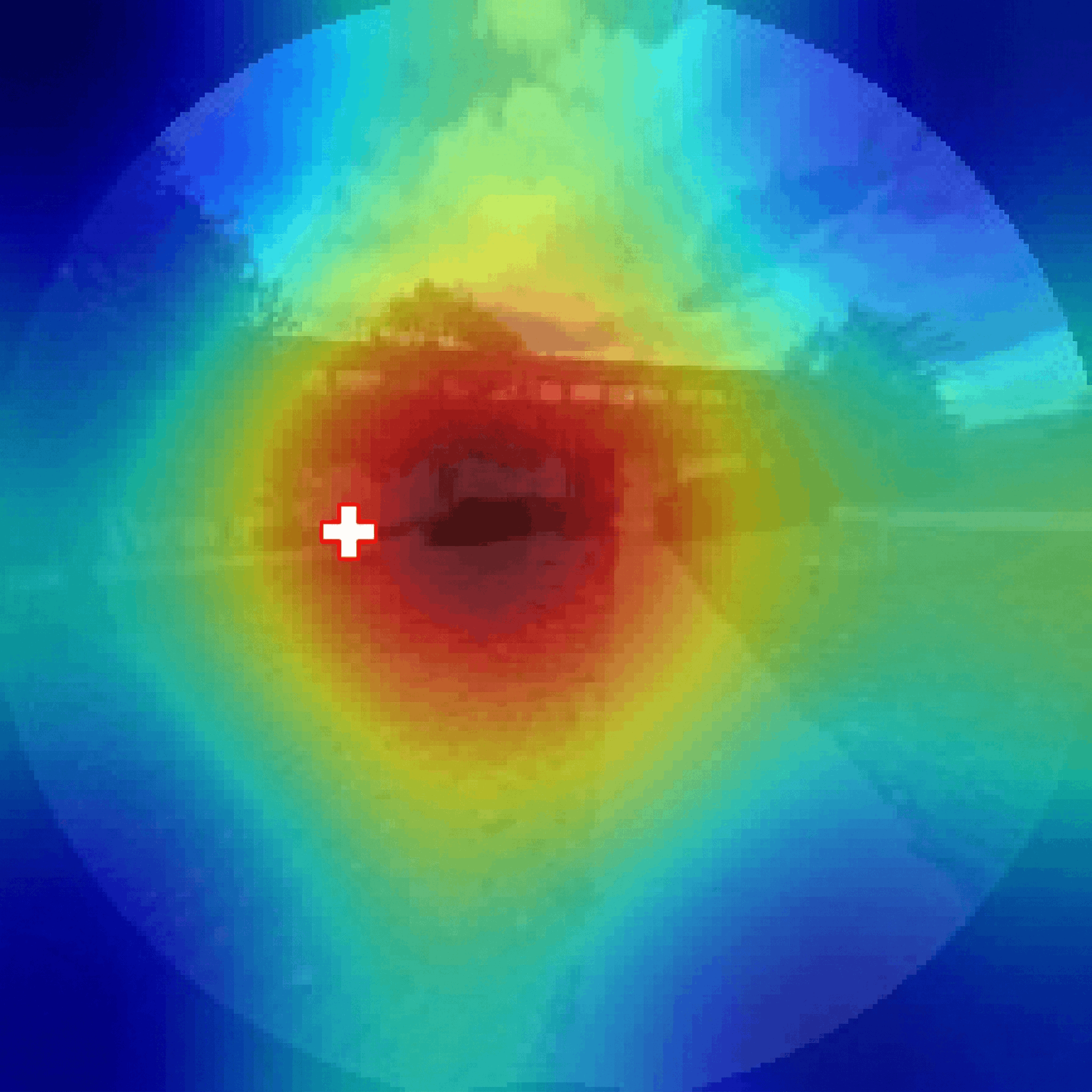} &
    \includegraphics[width=0.11\textwidth]{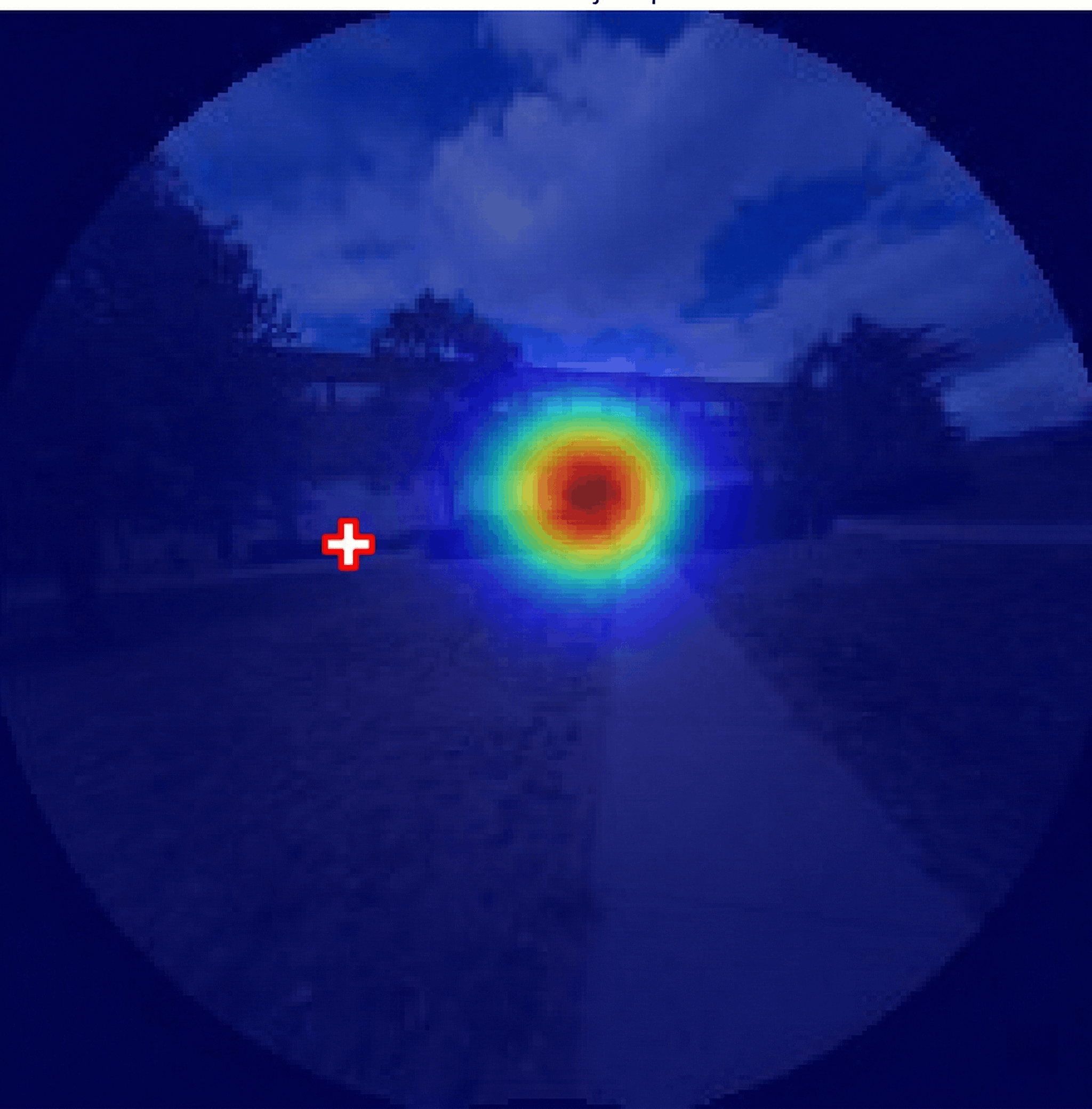}
    \\
    
    %=== Row 5 ===
    \includegraphics[width=0.11\textwidth]{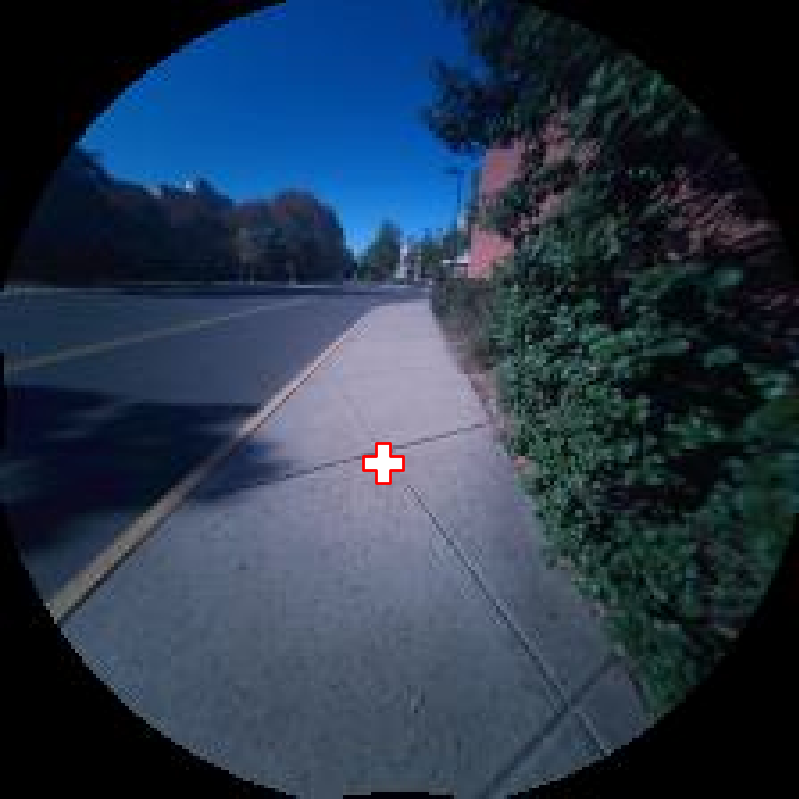} &
    \includegraphics[width=0.11\textwidth]{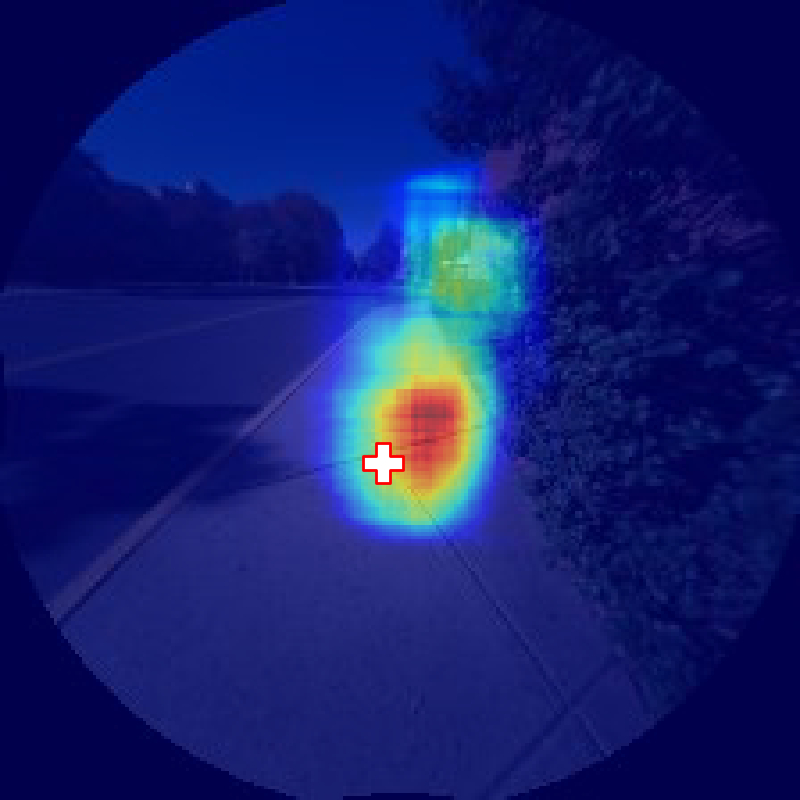} &
    \includegraphics[width=0.11\textwidth]{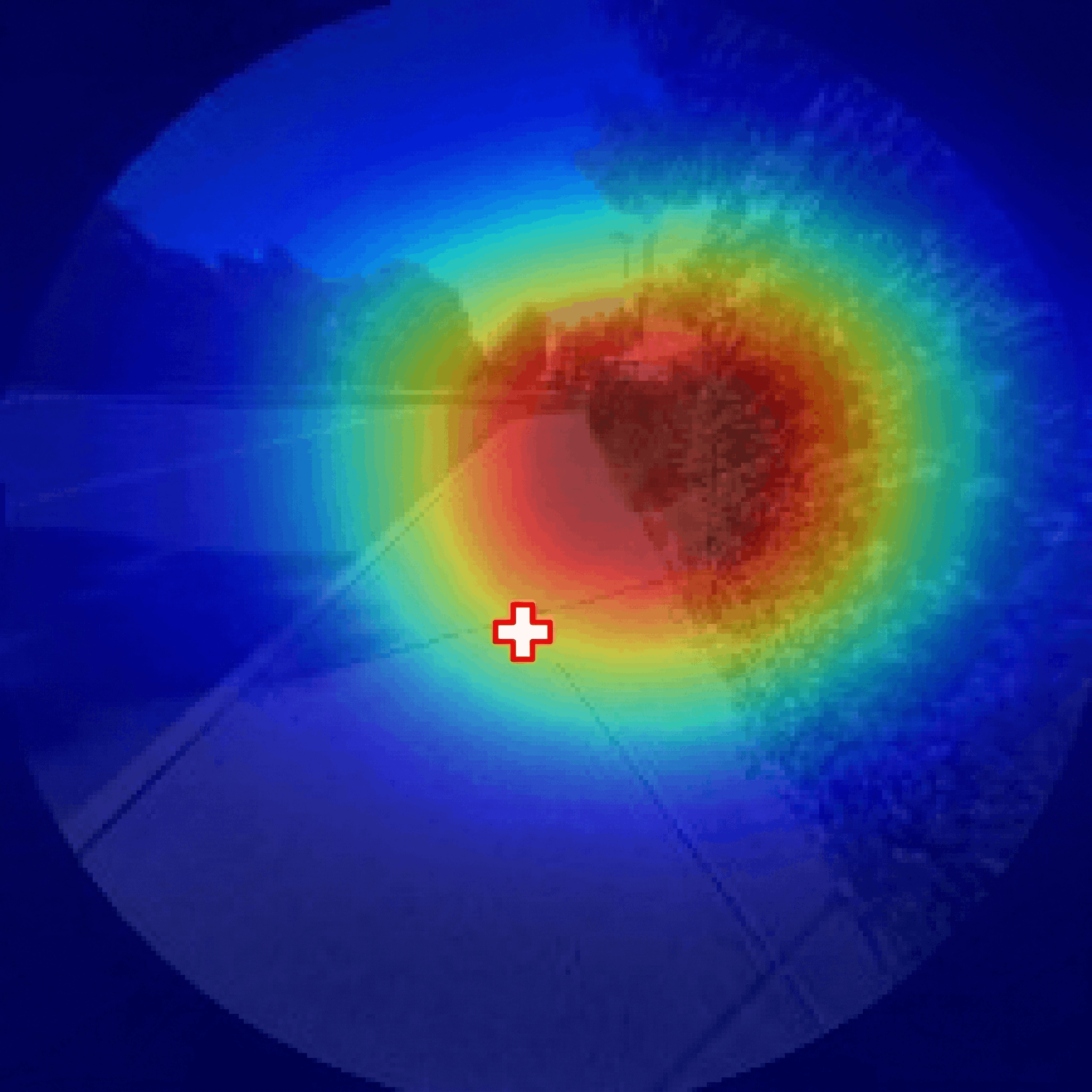} &
    \includegraphics[width=0.11\textwidth]{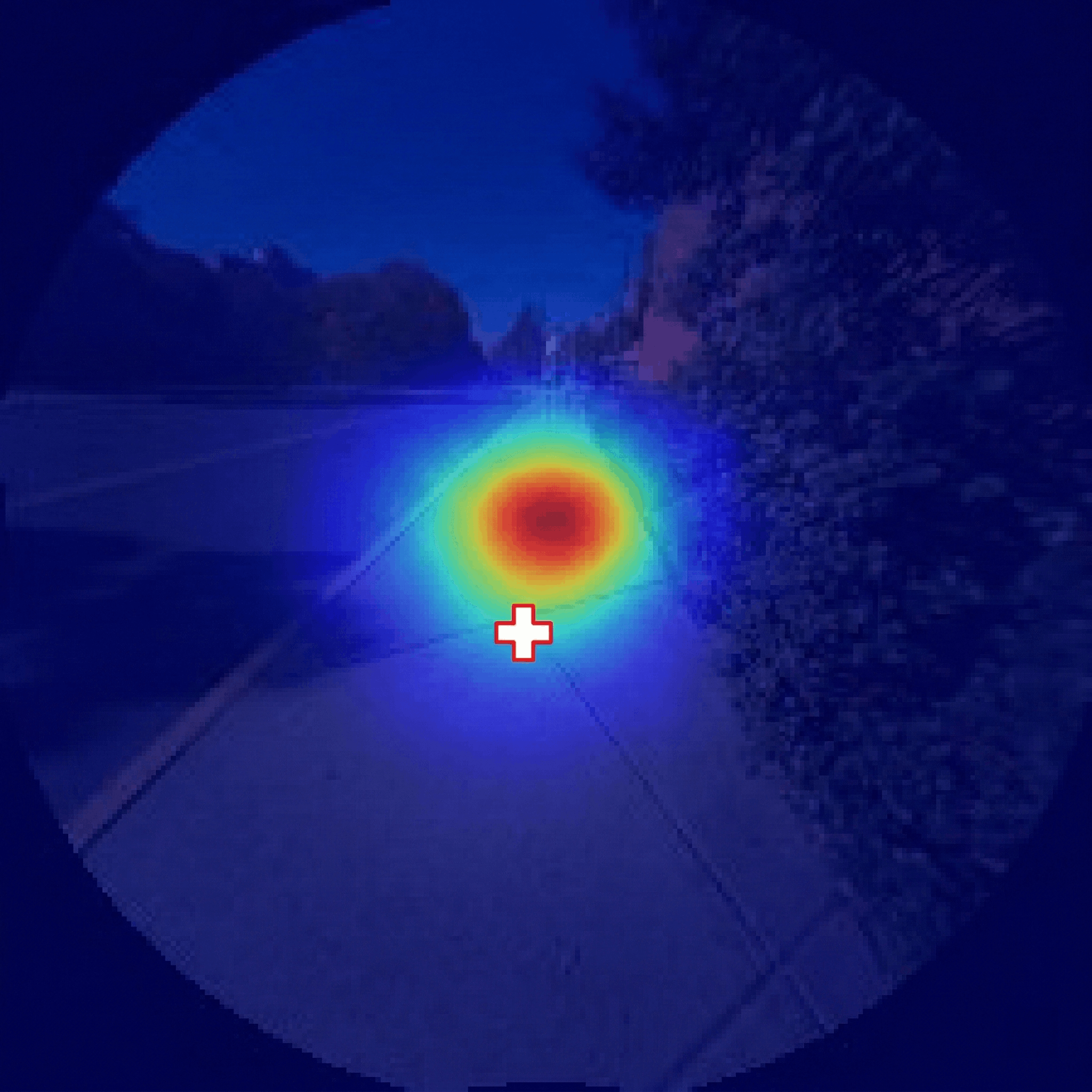} &
    \includegraphics[width=0.11\textwidth]{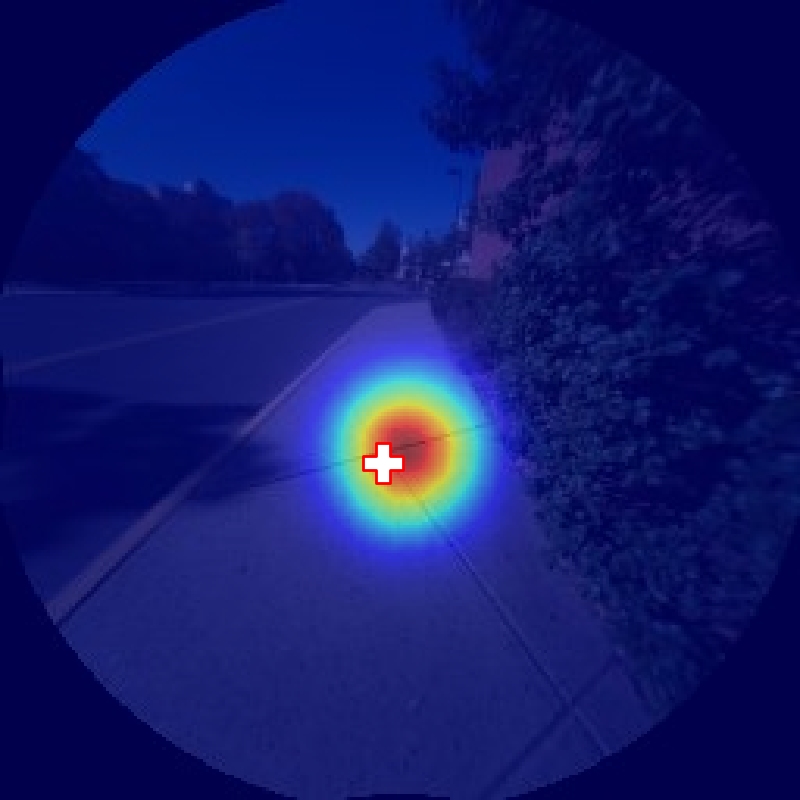} &
    \includegraphics[width=0.11\textwidth]{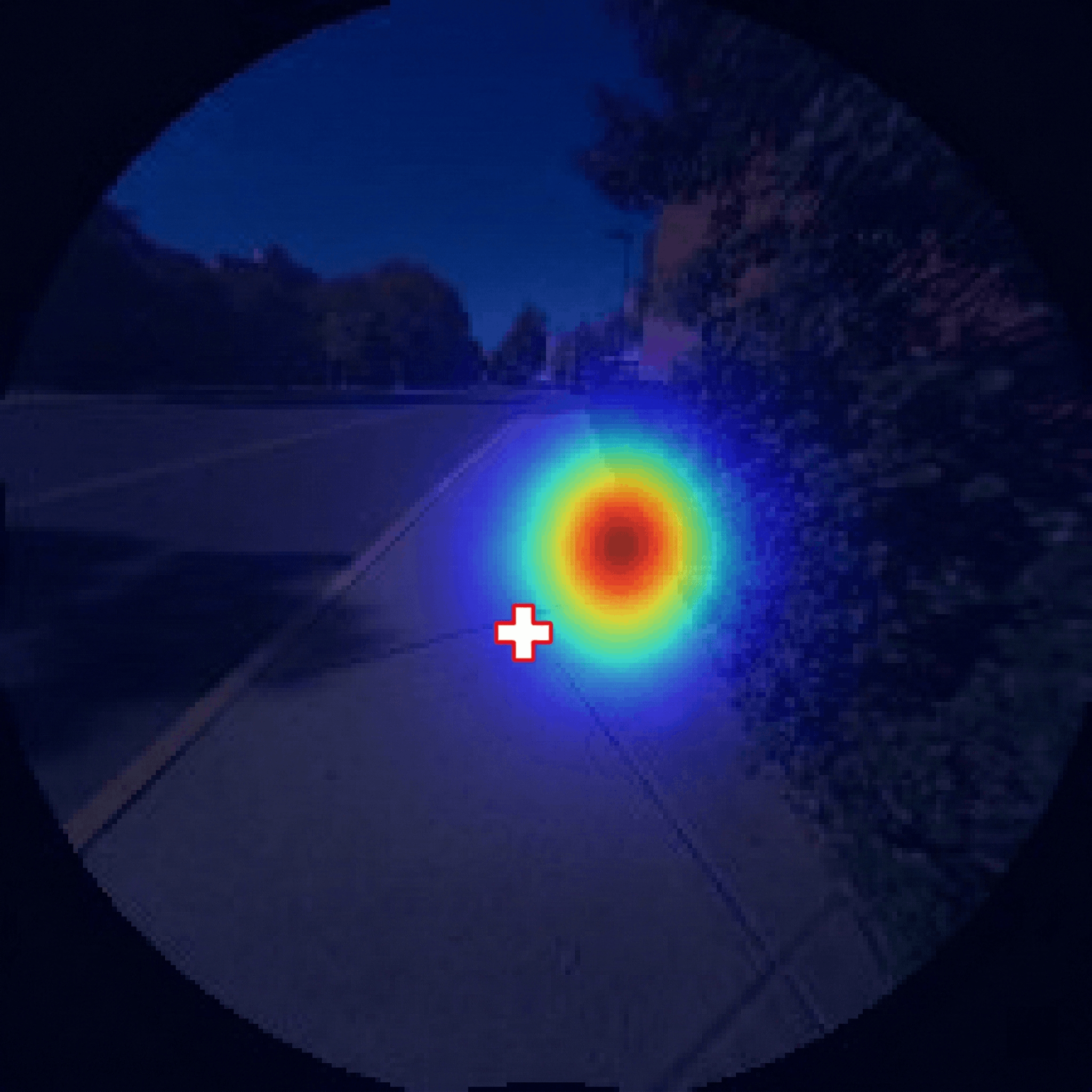} &
    \includegraphics[width=0.11\textwidth]{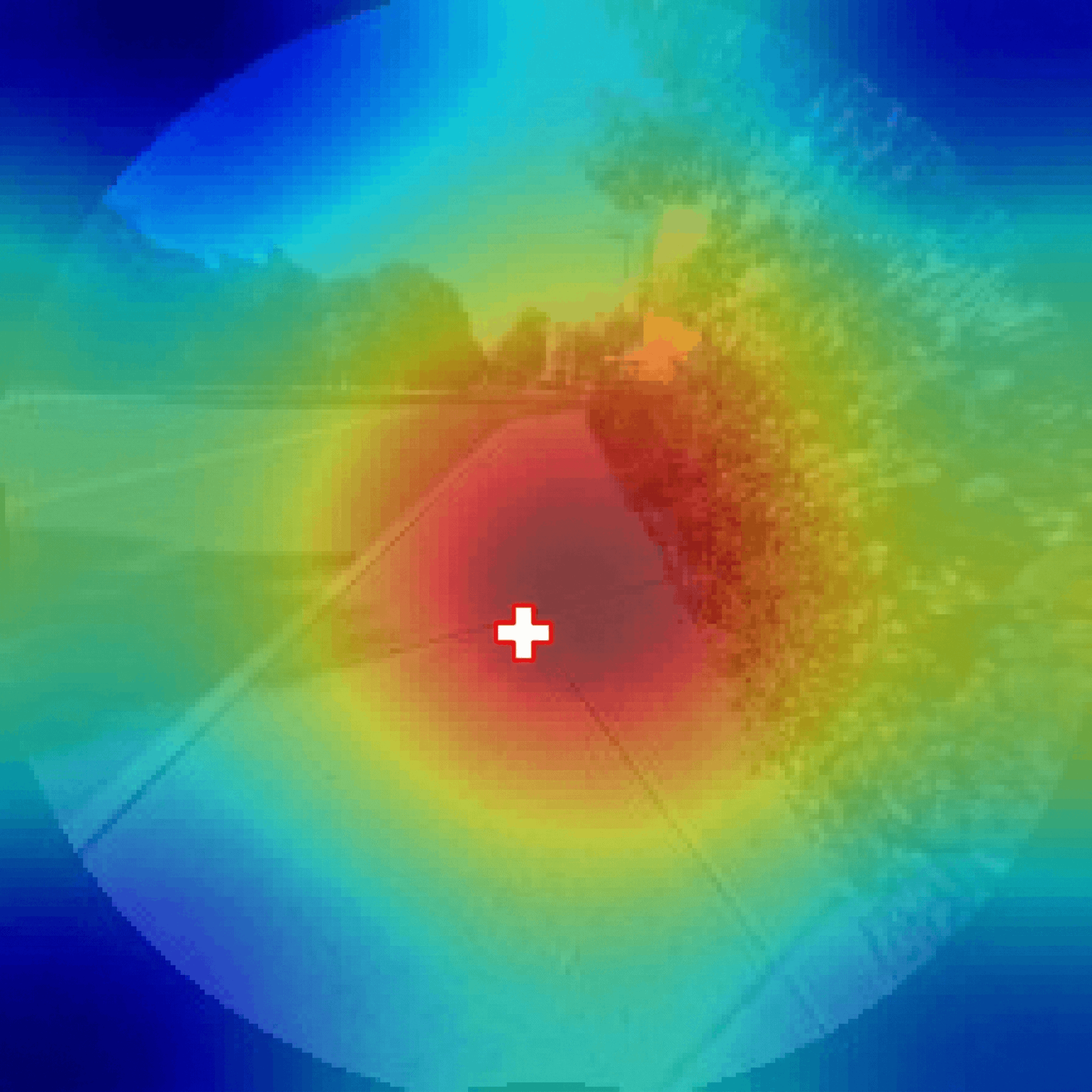} &
    \includegraphics[width=0.11\textwidth]{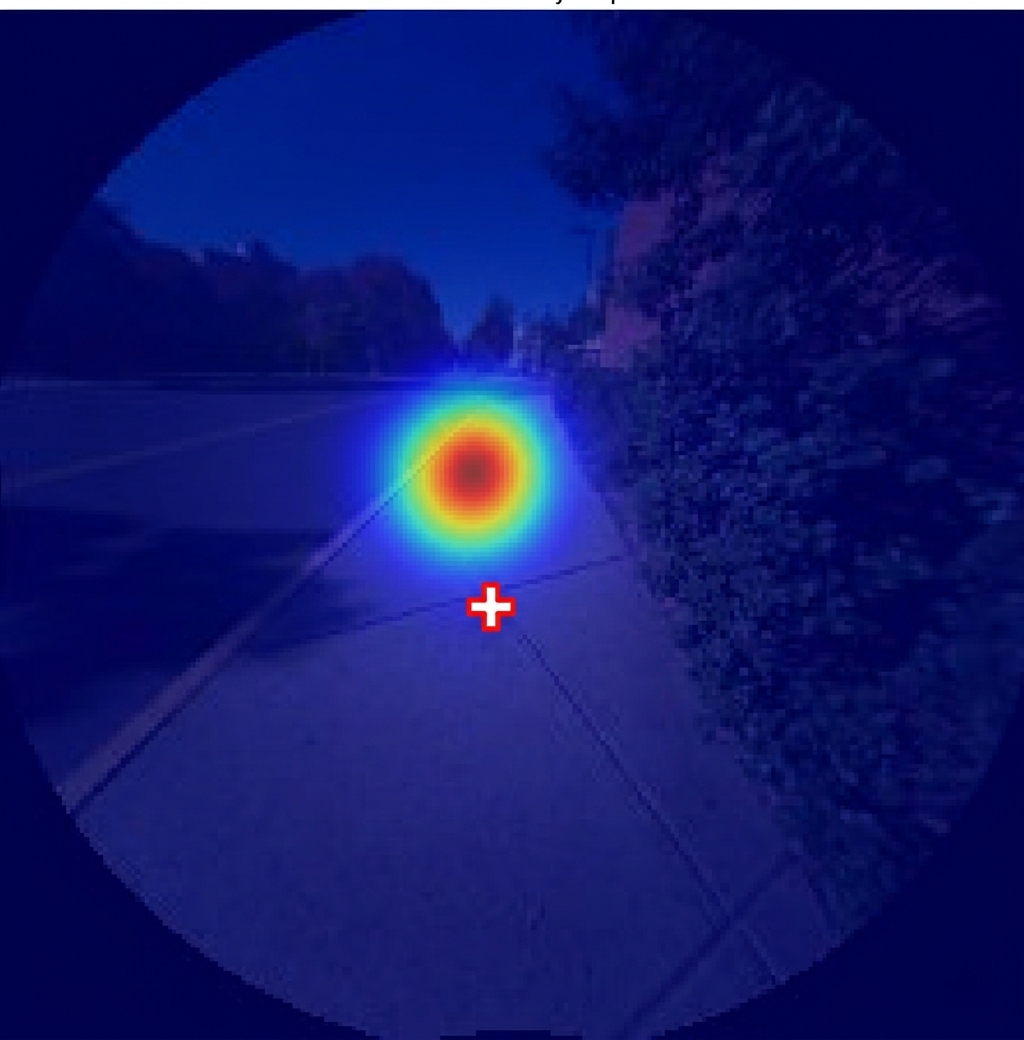}
    \\

    %=== Column labels ===
        {\scriptsize Query Frame} 
      & {\scriptsize\textbf{ECN (Ours)}}
      & {\scriptsize EML‑NET \cite{JIA20EML}}
      & {\scriptsize SUM \cite{Hosseini_2025_WACV}}
      & {\scriptsize GLC \cite{Lai2024}  }
      & {\scriptsize CV\_MM \cite{moskalenko2024aim}}
      & {\scriptsize DeepGazeIIE \cite{Linardos_2021_ICCV}}
      & {\scriptsize VistaHL \cite{moskalenko2024aim}}
      \\
  \end{tabular}

\caption{\textbf{Qualitative comparison of predicted gaze heat maps on sample frames.} For each scene (row), we show the query frame with the gaze point, the output from our proposed model, and outputs from six comparison methods. Our model consistently produces a more focal heatmap that is more accurately centered on the true gaze point. Note that the white $+$ icon denotes true gaze point in all images.}
  \label{fig:saliency-visualization}
\end{figure*}

\paragraph{Similarity (SIM):}SIM quantifies the degree of structural overlap between the predicted and ground truth saliency distributions. To compute this, both the predicted saliency map ($P$) and the ground truth fixation map ($Q$) are first normalized into probability mass functions (PMFs), where the sum of all pixel values in each map equals 1. $Q$ is the binary map divided by the count of fixations, and $P$ is the heatmap divided by the sum of all its values. 

\begin{equation}
\text{SIM} = \sum_{i} \min(P(i),Q(i))
\end{equation}

This calculation measures the intersection area of the two probability distributions. A score of 1 would mean a perfect match in the shape, location, and density of the two distributions, and 0 would mean no overlap. This metric rewards predictions that match the overall structure of the ground truth, regardless of absolute intensity.

\paragraph{Prior-Relative Weighting:} In our dataset, we observe the strong presence of a center bias. In order to better evaluate the relative performance of in the presence of this bias, we device a strategy for weighting the other performance metrics, de-emphasizing metrics when the ground truth is well explained by the dataset prior. We first obtain the dataset prior, calculated as the averaged the per-frame gaze heatmaps over the test set. The calculation of any metric is then weighted according to the following:

$$\text{Weight} = \text{JSD}(P\ \Vert\ \text{prior})$$

where JSD is the Jensen-Shannon divergence. JSD is similar to KLD, but is symmetric, and gives results bounded in the range of zero (if the distributions are identical) to one (if the distributions are entirely different). Additionally, the set of weights for all prediction \& label pairs are normalized such that they sum to one. When calculating metrics, this weighting effectively reduces the contribution from frames where the ground truth follows the dataset prior. This weighting is applied to all metrics in \cref{tab:weighted-metrics}.

\subsection{Quantitative Evaluation}\label{sec:comparisons}

As shown in \cref{tab:final-combined-leaderboard}, ECN proves to serve as a reasonable baseline for learned models, achieving competitive performance with a considerably smaller parameter count and reduced training costs. However, it does not achieve the best performance. While GLC \cite{Lai2024} achieves top performance by leveraging large-scale pretrained backbones, the Dataset and Center Priors remain remarkably competitive, often outperforming all methods except ECN and GLC. This confirms a dominant center bias in egocentric locomotion. To isolate performance from this bias, we use a weighting strategy that emphasizes frames where gaze deviates from the dataset prior. Under these weighted metrics, performance generally decreases, particularly for the priors (up to 18\% drop) and DeepGazeIIE \cite{Linardos_2021_ICCV}, revealing their heavy reliance on center-bias. ECN shows a more modest reduction, while GLC demonstrates superior robustness, actually improving in metrics like F1 by 9\%. These results underscore that while priors are strong baselines, models like GLC and ECN better capture active, off-axis attentional shifts.

\begin{table*}[t!]
  \centering
  \scriptsize
  \setlength{\tabcolsep}{3.8pt}
  \begin{tabular}{@{}lcccccccc@{}}
    \toprule
    Model 
      & AUC-J $\uparrow$ 
      & CC $\uparrow$ 
      & KLD $\downarrow$
      & SIM $\uparrow$
      & F1 $\uparrow$
      & Recall $\uparrow$
      & Prec. $\uparrow$
      & Params \\
    \midrule
    Dataset Prior 
    & 0.980 & 0.708 & 0.684 & 0.569 & 0.623 & 0.552 & 0.716 & - \\
    Center Prior                                         
    & 0.979 & 0.663 & 0.907 & 0.557 & 0.613 & 0.542 & 0.704 & - \\
    \midrule
    GLC \cite{Lai2024}                                   
    & 0.973 & 0.679 & 0.871 & 0.561 & 0.656 & 0.631 & 0.682 & 70.2M \\
    EML-NET \cite{JIA20EML}                
    & 0.984 & 0.482 & 1.029 & 0.418 & 0.587 & 0.671 & 0.522 & 47.2M \\
    SUM \cite{Hosseini_2025_WACV}          
    & 0.931 & 0.522 & 1.148 & 0.479 & 0.599 & 0.602 & 0.595 & 57.5M \\
    CV\_MM \cite{moskalenko2024aim}  
    & 0.971 & 0.720 & 0.962 & 0.612 & 0.695 & 0.703 & 0.687 & 420.5M \\
    VistaHL \cite{moskalenko2024aim}      
    & 0.973 & 0.669 & 1.027 & 0.420 & 0.593 & 0.601 & 0.585 & 187.7M \\
    DeepGazeIIE \cite{Linardos_2021_ICCV} 
    & 0.974 & 0.461 & 1.954 & 0.432 & 0.558 & 0.602 & 0.521 & 104M \\
    \midrule
    \textbf{ECN (Ours)}                  
    & 0.972 & 0.696 & 0.816 & 0.555 & 0.614 & 0.544 & 0.705 & 42.5M \\
    \bottomrule
\end{tabular}
  
\caption{Comprehensive performance comparison of methods trained on the EgoCampus dataset. After training on in-domain data, our ECN model achieves competitive results across both distribution-based and classification-based metrics while maintaining a significantly lower parameter count. However, GLC achieves best performance across most metrics.}

\label{tab:final-combined-leaderboard}
\end{table*}

\begin{table}[h!t]

  \centering
  \scriptsize
  \setlength{\tabcolsep}{3.3pt}
  \begin{tabular}{@{}lcccccccc@{}}
    \toprule
    Model 
      & AUC-J $\uparrow$ 
      & CC $\uparrow$ 
      & KLD $\downarrow$
      & SIM $\uparrow$
      & F1 $\uparrow$
      & Recall $\uparrow$
      & Prec. $\uparrow$ \\
    \midrule
    Dataset Prior 
    & 0.9818 & 0.5913 & 1.0125 & 0.4840 & 0.5227 & 0.4630 & 0.5227 \\
    Center Prior                                         
    & 0.9811 & 0.5293 & 1.5694 & 0.4597 & 0.5119 & 0.4529 & 0.5119 \\
    \midrule
    GLC \cite{Lai2024}                                   
    & 0.9884 & 0.7669 & 0.4818 & 0.6191 & 0.7157 & 0.6894 & 0.7157 \\
    EML-NET \cite{JIA20EML}                
    & 0.9821 & 0.4376 & 1.1728 & 0.3794 & 0.5505 & 0.6323 & 0.4874 \\
    SUM \cite{Hosseini_2025_WACV}          
    & 0.9310 & 0.4804 & 1.7571 & 0.4131 & 0.5422 & 0.5497 & 0.5348 \\
    CV\_MM \cite{moskalenko2024aim}  
    & 0.9686 & 0.6591 & 1.1510 & 0.5346 & 0.6339 & 0.6327 & 0.6352 \\
    VistaHL \cite{moskalenko2024aim}      
    & 0.9675 & 0.6081 & 1.1292 & 0.3748 & 0.5509 & 0.5655 & 0.5370 \\
    DeepGazeIIE \cite{Linardos_2021_ICCV} 
    & 0.9741 & 0.3962 & 2.4747 & 0.3733 & 0.4565 & 0.5004 & 0.4197 \\
    \midrule
    \textbf{ECN (Ours)}                  
    & 0.9740 & 0.5928 & 1.1990 & 0.4784 & 0.5296 & 0.4691 & 0.5296 \\
    \bottomrule
  \end{tabular}

\caption{Performance comparisons with \textbf{metrics weighted relative to dataset prior} as described in \cref{sec:metrics}. Models that overfit and heavily rely on the the center prior are effectively penalized, while models that make more diverse predictions retain or even gain performance, relative to \cref{tab:final-combined-leaderboard}.}
\label{tab:weighted-metrics}
\end{table}

\subsection{Qualitative Results}\label{sec:qualitative}

Qualitative results of different methods on our dataset is shown in \cref{fig:saliency-visualization}. EML-Net \cite{JIA20EML} and DeepGazeIIE \cite{Linardos_2021_ICCV} tend to overestimate the likely gaze area.  ECN learns a strong contribution from the center bias, often predicting the center of the image, which tends to be the direction of the subject's movement during recording. When the subject walks past other pedestrians, other models often predict that other peoples' faces is the most likely place to look, but in the setting of egocentric locomotion, this is often not the case. Other strong performing models (SUM, GLC, CV\_MM) make conservative estimates near the center, while still occasionally focusing on pedestrians or features in the distance.
\vspace{-0.3 cm}

\section{Conclusion}

\vspace{-0.1cm}

% summary of work
In this work, we introduce EgoCampus, a novel multimodal egocentric dataset designed to facilitate the study of human attention during real-world navigation. By capturing synchronized video, eye gaze, and inertial data from numerous participants traversing shared outdoor routes, EgoCampus provides a unique resource for building and evaluating future visual attention models. We present EgoCampusNet, which fuses spatio-temporal features from a video encoder backbone with image features learned from the individual query frame. Our experiments demonstrate that this approach sets an efficient benchmark for our dataset.
% applications of work
The findings from this research have significant implications for embodied AI and human-robot interaction. The EgoCampus dataset opens avenues for investigating more complex pedestrian behavior, pushing the frontier of environment-aware attention modeling. 
% concurrent with Jackal

The EgoCampus dataset will be made publicly available and has a {\bf companion dataset}  YOPO-Campus \cite{Anonymous2025}, where the same campus paths ($6$ km) were traversed by a tele-operated Clearpath Jackal robot. While YOPO-Campus shows the paths from a robot view, EgoCampus shows the paths from a pedestrian view and capture eye gaze. Together, both datasets provide an important data resource for future studies of navigation in human-robot systems.

\section*{Acknowledgments}
 This work was supported by the NSF-NRT grant: Socially Cognizant Robotics for a Technology Enhanced Society (SOCRATES), No.\ 2021628 and NSF CNS Reality-Aware Networks No.\ 1901355. Thank you to Meta for allowing us to use the Project Aria glasses.

{
    \small
    \bibliographystyle{ieeenat_fullname}
    \bibliography{main}
}

\end{document}